\documentclass[10pt,twocolumn,letterpaper]{article}
\usepackage{etex}
\usepackage{iccv}
\usepackage{times}
\usepackage{epsfig}
\usepackage{graphicx}
\graphicspath{{images/}}
\usepackage{amsmath}
\usepackage{amssymb}
\usepackage{amsfonts}
\usepackage[tight]{units}
\usepackage[caption=false,font=footnotesize]{subfig}
\usepackage{multirow}
\usepackage{booktabs}
\usepackage{pgfplots}
\usepackage[normalem]{ulem}
\usepackage{footnote}
\usepackage{xspace}
\usepackage{tikz}
\pgfplotsset{compat=newest}
\usetikzlibrary{external} 

\usepackage{abbrevs}
\newcommand{\subsequentFont}[1]{#1}
\newcommand{\fMRF}{Markov Random Field}        \newabbrev\MRF{\subsequentFont{\fMRF} (\textit{MRF})}[\textit{MRF}]
\newcommand{\fTOF}{Time of Flight}             \newabbrev\TOF{\subsequentFont{\fTOF} (\textit{ToF})}[\textit{ToF}] \newabbrev\uTOF{\fTOF}
\newcommand{\fTGV}{Total Generalized Variation}\newabbrev\TGV{\subsequentFont{\fTGV} (\textit{TGV})}[\textit{TGV}]
\newcommand{\fGPU}{Graphical Processing Unit}\newabbrev\GPU{\subsequentFont{\fGPU} (\textit{GPU})}[\textit{GPU}]
\newcommand{\fLF}{Legendre Fenchel transform} \newabbrev\LF{\subsequentFont{\fLF} (\textit{LF})}[\textit{LF}]
\newcommand{\fSR}{upsampling} \newabbrev\SR{\fSR} \newabbrev\uSR{Upsampling}
\newcommand{\fRMSE}{root mean squared error} \newabbrev\RMSE{\subsequentFont{\fRMSE} (\textit{RMSE})}[\textit{RMSE}]
\newcommand{\fSNR}{signal to noise ratio} \newabbrev\SNR{\subsequentFont{\fSNR} (\textit{SNR})}[\textit{SNR}]
 \newabbrev\SOTA{state of the art} \newabbrev\uSOTA{State of the art}

\iccvfinalcopy 

\def\iccvPaperID{0122} 

\ificcvfinal\fi

\usepackage[textwidth=1.8cm,figwidth=\columnwidth,colorinlistoftodos,disable]{todonotes}

\newcounter{mycomment} 

\newcommand{\mat}[1]{\ensuremath{#1}}
\renewcommand{\vec}[1]{\ensuremath{{#1}}}

\DeclareMathOperator*{\argmin}{\arg\!\min}
\newcommand*{\Ds}{\ensuremath{\mat{D}_\mathrm S}\xspace}
\newcommand*{\Ih}{\ensuremath{\mat{I}_\mathrm H}\xspace}
\newcommand*{\Il}{\ensuremath{\mat{I}_\mathrm L}\xspace}
\newcommand*{\Dh}{\ensuremath{\mat{D}_\mathrm H}\xspace}
\newcommand*{\Dl}{\ensuremath{\mat{D}_\mathrm L}\xspace}
\newcommand*{\Dsij}{\ensuremath{\mat{D}_{\mathrm S} \, i,j}\xspace}
\newcommand*{\Dlij}{\ensuremath{\mat{D}_{\mathrm L} \, i,j}\xspace}

\usepackage[pagebackref=false,breaklinks=true,colorlinks,bookmarks=false, plainpages=false,naturalnames=false,hypertexnames=true]{hyperref}

\begin{document}
\title{Image Guided Depth Upsampling using Anisotropic Total Generalized Variation}

\author{David Ferstl, Christian Reinbacher, Rene Ranftl, Matthias R\"uther and Horst Bischof\\
Graz University of Technology\\
Institute for Computer Graphics and Vision\\
Inffeldgasse 16, 8010 Graz, AUSTRIA\\
{\tt\small \{ferstl,reinbacher,ranftl,ruether,bischof\}@icg.tugraz.at} 
}

\maketitle

\begin{abstract}
In this work we present a novel method for the challenging problem of depth image upsampling. Modern depth cameras such as Kinect or \uTOF cameras deliver dense, high quality depth measurements but are limited in their lateral resolution. To overcome this limitation we formulate a convex optimization problem using higher order regularization for depth image upsampling. In this optimization an anisotropic diffusion tensor, calculated from a high resolution intensity image, is used to guide the upsampling. We derive a numerical algorithm based on a primal-dual formulation that is efficiently parallelized and runs at multiple frames per second. We show that this novel upsampling clearly outperforms \SOTA approaches in terms of speed and accuracy on the widely used Middlebury 2007 datasets. Furthermore, we introduce novel datasets with highly accurate groundtruth, which, for the first time, enable to benchmark depth upsampling methods using real sensor data.

\end{abstract}
\graphicspath{{images/}}

\ResetAbbrevs{All}
\section{Introduction}
Accurate, high resolution depth sensing is a fundamental challenge in computer vision. It is used in a variety of different applications including object reconstruction, robotic navigation and automotive driver assistance. Traditional computer vision approaches calculate the scene depth through computational exhaustive stereo calculations or expensive laser range measurements. 

Recently, \TOF range sensors became a popular alternative for dense depth sensing. 
A per-pixel depth is measured actively through the runtime of light. The measurement is independent from scene texture and largely independent from environmental lighting conditions. It delivers a dense depth map even at very close ranges \cite{Lange2000, Schmidt2011}. No additional calculations are necessary, which results in depth measurements at high frame rates.
Recently, \TOF sensors have become affordable in the mass market and a small packet size and a low energy consumption make them applicable in mobile devices. However, their main disadvantages are a low resolution caused by chip size limitations and acquisition noise due to limited active illumination energy.
\begin{figure}
  \begin{center}
  \subfloat[Low resolution depth]{ 
    \includegraphics*[height=0.34\linewidth]{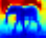}
  }
  \subfloat[High resolution intensity]{ 
    \includegraphics*[height=0.34\linewidth]{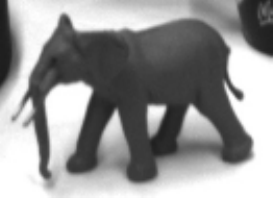}
  } \\
  \subfloat[High resolution depth upsampling result]{ 
    \includegraphics*[height=0.34\linewidth]{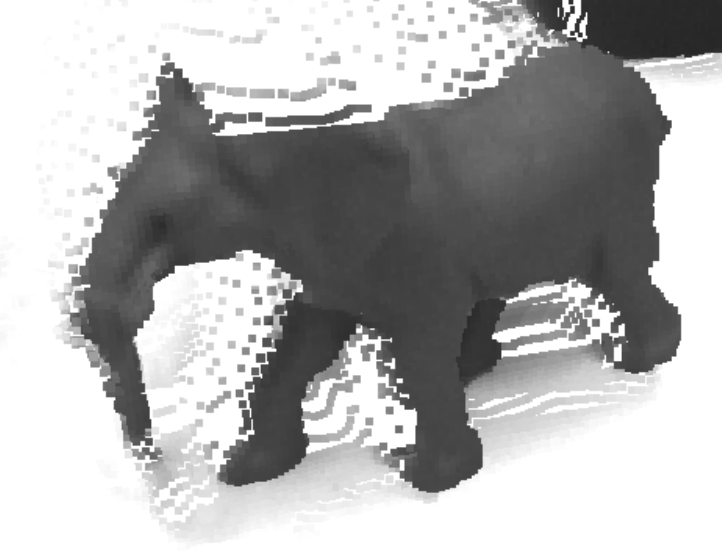} \hspace{1pt}
    \includegraphics*[height=0.34\linewidth]{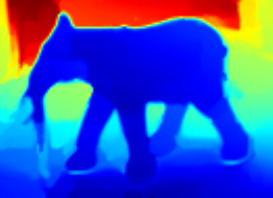}
  }
\end{center}
  \caption{Upsampling of a low resolution depth image (a) using an additional high resolution intensity image (b) through image guided anisotropic Total Generalized Variation (c). Depth maps are color coded for better visualization.}
\label{fig:intro}
\end{figure}

In this work, we propose a method to drastically increase the lateral measurement resolution by a novel depth map 
upsampling approach, as shown in Figure \ref{fig:intro}. To increase both, quality and resolution, we add information 
from a high resolution intensity camera in a variational optimization framework. We build on the observation that 
textural edges are more likely to appear at high depth discontinuities, whereas homogeneous textured regions correspond 
to homogeneous surface parts \cite{Torralba2003}. Fusing both, low resolution but very robust depth and high resolution 
intensity in a spatial sense, results in a dense depth map with increased lateral resolution and visual quality.

We formulate the \SR as a convex optimization problem \cite{Chambolle2011,Esser2010}. The energy is composed of two terms. First, the data term forces the solution to be similar to the input depth measurements. Second, the higher order regularization term enforces a piecewise affine solution, preserving sharp edges according to the texture, while compensating acquisition noise. This term is modeled as a second order \TGV regularization and is weighted according to the intensity image texture by an anisotropic diffusion tensor. 

The main contributions of this work are two-fold: (1) We propose a novel method for fast depth image \SR by combining a low resolution depth image with high resolution texture information in a variational energy optimization framework. The employed higher order regularization is well suited to model the image acquisition process of modern depth cameras and leads to an improved quality of the upsampled depth maps, compared to \SOTA methods.
(2) We propose benchmarking datasets that enable a quantitative comparison of depth image \SR methods providing real 
\TOF and intensity camera acquisitions together with a highly accurate groundtruth measurement. To encourage further 
comparison and future work, these novel datasets and MATLAB code of our method are available at our website 
\footnote{\url{http://rvlab.icg.tugraz.at/tofmark}}. 

In our experiments we demonstrate the \SR quality by a numerical and visual comparison on synthetic and real benchmarking datasets. Compared to \SOTA methods, our method is superior in terms of speed and accuracy on all test sets.
\ResetAbbrevs{All}
\section{Related Work}
There are many ways to increase the resolution and the accuracy of depth measurements. In general, they can be separated 
in three main classes: (1) fusion of multiple depth sensors, (2) temporal and spatial fusion and (3) \SR by combining 
depth and intensity sensors.
\paragraph*{Multiple Depth Sensor Fusion}
Recent works addressed the fusion of different depth sensing techniques to increase resolution and quality. Gudmundsson 
\etal. \cite{Gudmundsson2008} presented a method for stereo and \TOF depth map fusion in a dynamic programming approach. 
Similar work has been proposed by Zhu \etal \cite{Zhu2011} using an accurate depth calibration and fusing the 
measurements in a \MRF framework. Additionally to this spatial fusion also a temporal fusion was performed by measuring 
the frame-to-frame displacement acquired with high speed intensity cameras.
\paragraph*{Temporal and Spatial \uSR}
A common way to improve the resolution and quality of depth information is to fuse multiple depth measurements into one depth map. Schuon \etal \cite{Schuon2009} proposed a method to fuse \TOF acquisitions of slightly moved viewpoints. It uses a bilateral regularization in a \MRF optimization framework incorporating also the \TOF sensor characteristics. Based on this work, Cui \etal \cite{Cui2010} used a set of fused depth maps with larger displacements. To create whole volumes of depth data Newcombe \etal \cite{Newcombe2011} proposed a method for simultaneous camera localization and depth fusion in real time. 
\vspace*{-1.5mm}
\paragraph*{Depth \uSR through Intensity Information}
This class of approaches uses additional intensity information as depth cue for image upsampling. Yang \etal \cite{Yang2007} used bilateral filtering of a depth cost volume and a RGB image in an iterative refinement process. Chan \etal \cite{Chan2008} used a noise aware joint bilateral filter to increase the resolution and to reduce depth map errors at multiple frames per second. Diebel and Thrun \cite{Diebel2006} performed an upsampling using a \MRF formulation, where the smoothness term is weighted according to texture derivatives. A more complex approach was proposed by Park \etal \cite{Park2011}. They used a combination of different weighting terms of a least squares optimization including segmentation, image gradients, edge saliency and non-local means for depth upsampling. The combination of intensity and depth data in a Bayesian Framework was proposed by Li \etal \cite{Li2012}.
\vspace*{-1.5mm}
\paragraph*{Discussion}
While the methods for multiple sensor fusion deliver accurate depth results, their quality relies on high calibration effort. Further, most sensor fusion techniques have to calculate a depth map from passive stereo in a preprocessing step before the actual fusion is able to start. Contrary, temporal and spatial fusion approaches rely on multiple acquisitions from a single depth sensor. The major drawback of these methods is that changing environments during these acquisitions will harm the fusion result.

To overcome these limitations, we chose the combination of a low resolution depth and a high resolution intensity sensor to increase the natural depth sensor resolution. The upsampling is calculated on a per image basis without the need for complex preprocessing. Existing approaches, such as \cite{Chan2008, Yang2007}, calculate this depth upsampling by a bilateral filtering. While bilateral filtering techniques can operate at high frame rates they have a drawback in oversmoothing fine details. In contrast, our method builds on the success of recently introduced upsampling methods using \MRF and least squares optimization \cite{Diebel2006,Park2011}. Unlike them, our approach incorporates a higher order regularization, which avoids surface flattening. Furthermore, we use an anisotropic diffusion tensor based on the intensity image. This tensor not only weights the depth gradient but also orients the gradient direction during the optimization process.

\ResetAbbrevs{All}
\section{Method}
Our \SR approach generates a high quality and high resolution depth map \Dh out of a high resolution intensity image \Ih and a low resolution and noisy depth map \Dl, where $\Ih, \Dh \colon \Omega_{\mathrm H} \subseteq \mathbb{R}^{2}$ and $\Dl\colon \Omega_{\mathrm L} \subseteq \mathbb{R}^{2}$. The methodology of this approach can be divided into three main areas: (1) Registering the low-resolution depth measurements and the high resolution intensity information in one common coordinate system (Section~\ref{ssec:calibration}), (2) formulating the depth \SR problem into a convex energy functional (Section~\ref{ssec:energy_model}), and (3) solving the optimization problem with a first-order primal-dual optimization scheme (Section~\ref{ssec:optimization}).

\subsection{Depth Mapping} \label{ssec:calibration}
\begin{figure}[htbp]
   \centering
   \includegraphics[width=\columnwidth]{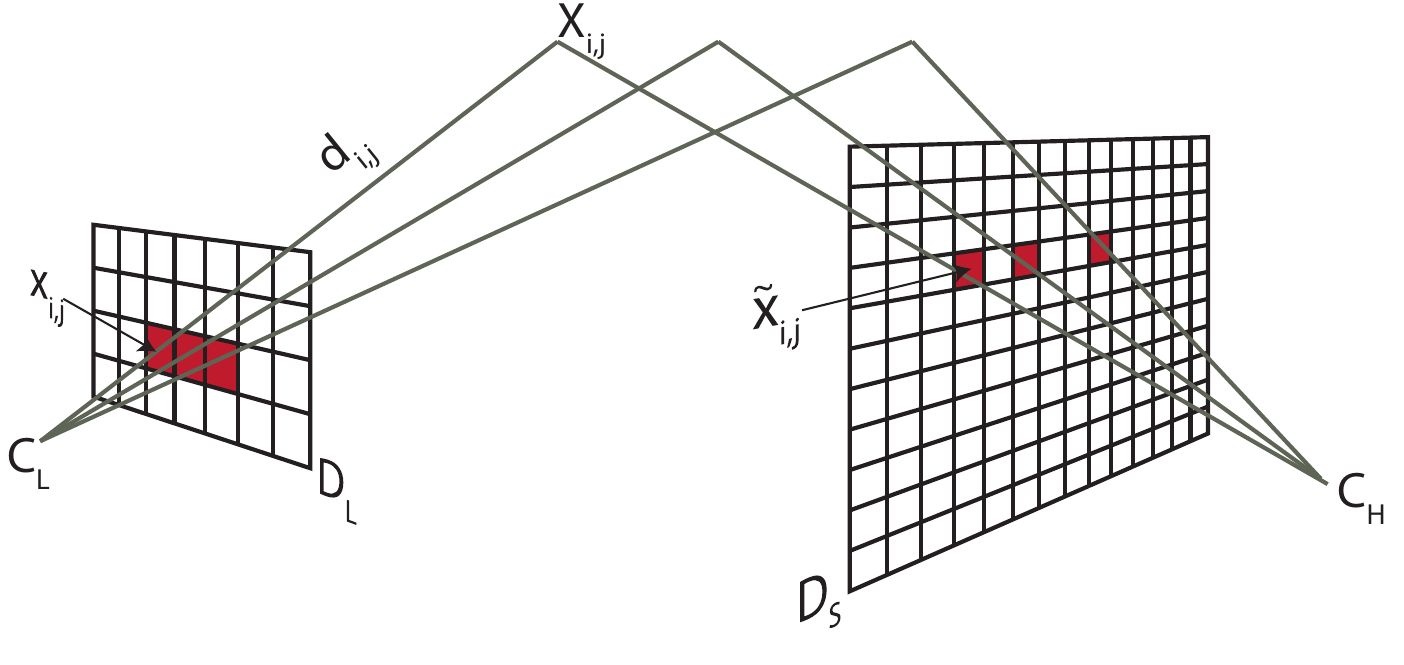}
    \caption{Projection from a low resolution depth map \Dl to a high resolution sparse depth map \Ds in the intensity camera coordinate system.}
   \label{fig:upsampling}
\end{figure}

Since the low resolution depth map \Dl and the high resolution intensity image \Ih stem from different cameras, a mapping can only be established when intrinsic and extrinsic parameters are known (see Section~\ref{ssec:benchmark}). In our setup we define the intensity camera as the world coordinate center. Each depth measurement $d_{i,j}$ at pixel position $\vec x_{i,j} = [i, j, 1]^{\mathrm T}$ is projected into the high resolution intensity image space $\Omega_{\mathrm H}$. This projection is calculated as
\begin{align}
\begin{split}
  \vec{X}_{i,j} &= \vec{C}_\mathrm L + d_{i,j} \frac{ \mat{P}_{\mathrm L}^{\dagger} \vec{x}_{i,j}}{\|\mat{P}_{\mathrm L}^{\dagger} \vec{x}_{i,j} \|} \\
  \tilde{\vec{x}}_{i,j} &= \mat{P}_{\mathrm H}\vec{X}_{i,j} \quad \forall i,j \in \Omega_{\mathrm L},
\label{equ:2d23d}
\end{split}
\end{align}
where $\mat{P}_{\mathrm L}^{\dagger}$ is the pseudoinverse of the depth camera projection matrix, $\vec{C}_\mathrm L$ the camera center and $\vec{X}_{i,j}$ the 3D point. Each 3D point is back projected by multiplication with the projection matrix of the intensity camera $\mat{P}_{\mathrm H}$. Hence, we get a projected depth image \Ds consisting of a sparse set of base depth points at position $\tilde{\vec{x}}_{i,j}$ in the intensity image space $\Omega_\mathrm H$ where the depth value is given by the distance to the 3D point $\vec{X}_{i,j}$ (see Figure~\ref{fig:upsampling}).

Although, one low resolution sensor pixel $\Dlij$ measures the average depth of multiple pixels in the high resolution space we only project it to one central pixel $\Dsij$ at position $\tilde{\vec{x}}_{i,j}$. 
Therewith, we minimize the error which can occur due to this averaging in the high resolution space. Through the regularization term, introduced in Section~\ref{ssec:energy_model}, the area between the projected depth pixels is implicitly interpolated.

\subsection{Depth Image \uSR}\label{ssec:energy_model}
Our \SR method increases the resolution of measured depth data from a low resolution depth sensor by adding edge cues from a high resolution intensity image. To be able to use both information, we map the depth measurements to the intensity camera coordinate system as described in Section~\ref{ssec:calibration}. With this mapping we get a depth map \Ds of a sparse set of base depth measurements from the low resolution depth sensor.

The high resolution depth map \Dh is given by
\begin{align}
\Dh = \argmin_u \left\lbrace 
	  G(u, \Ds) + \alpha F (u)
 \right\rbrace.
  \label{equ:energy_general}
\end{align}
This formulation is composed of the data term $G(u,\Ds)$ that measures the fidelity of the argument $u$ to the input depth measurements \Ds and the regularization term $F(u)$ that reflects prior knowledge of the smoothness of our solution. $F$ and $G$ are convex lower semi-continuous functions. The scalar $\alpha$ is used to balance the relative weight between the data and the regularization.

The data term in our energy model is designed to ensure a data consistency to the base depth points \Ds from the depth camera. 
Additionally, we allow to weight the depth measurements with a weighting operator $w = [0, 1] \in 
\mathbb{R}^{\Omega_H}$, which is zero at unmapped image points and between zero and one on the base points according to 
some application specific confidence. Hence, the data term results in
\begin{align}
    G(u,\Ds) = \int_{\Omega_H} w|(u - \Ds)|^{2} \,\mathrm dx,
  \label{equ:data_term}
\end{align}
which penalizes deviations of the resulting depth from the measured depth.
The regularization term has to meet the challenges of producing a high resolution depth map out of a sparse set of depth points. Most currently utilized regularization terms are based on the first order smoothness assumption \cite{Rudin1992}, \eg the Total Variation semi norm, which results in $F(u) = \|\nabla u \|_1$. While the simple model with L1 norm is well suited for intensity image denoising, it has a disadvantage when used for range data regularization. Through its gradient penalization it favors constant solutions. This prevents the depth map to become a piecewise smooth surface, resulting in piecewise fronto parallel depth reconstructions.

Hence, we use a more generalized regularization model namely the \TGV introduced by Bredies \etal \cite{Bredies2010}. 
The \TGV is composed of polynomials of arbitrary order, which allows to reconstruct piecewise polynomial functions. An 
order of $k$ favors solutions composed of polynomials of order $k-1$. For depth upsampling, it turns out that the second 
order \TGV is sufficient, since most objects can be well approximated by piecewise affine surfaces. The primal 
definition of the second order \TGV is formulated as
\begin{align}
\begin{split}
  \mathit{TGV}^{2}_{\alpha} = \min_{v} \left\lbrace \alpha_{1} \int_{\Omega} \mathrm |\nabla u - v| \,\mathrm dx + \alpha_{0} \int_{\Omega} \mathrm |\nabla v| \,\mathrm dx \right\rbrace
\end{split}
  \label{equ:energy_tgv_2},
\end{align}
where the scalars $\alpha_0$ and $\alpha_1$ are used to weight each order. Because the \TGV regularizer is convex it allows to compute a globally optimal solution.

Assuming that texture edges most likely correspond to depth discontinuities, we use the high resolution intensity data to produce a more accurate \SR result. Henceforth, we include an anisotropic diffusion tensor $T^{\frac{1}{2}} $. 
This tensor is calculated by
\begin{align}
  T^{\frac{1}{2}} = \exp \left( - \beta \left|\nabla \Ih \right|^{\gamma} \right) n n^{\mathrm T} + n^{\bot}n^{\bot \mathrm T}
  \label{equ:anisoTensor},
\end{align}
where $n$ is the normalized direction of the image gradient $n=\frac{\nabla \Ih}{|\nabla \Ih|}$, $n^{\bot}$ is the normal vector to the gradient and the scalars $\beta$, $\gamma$ adjust the magnitude and the sharpness of the tensor. The anisotropic diffusion tensor not only weights the first order depth gradient but also orients the gradient direction during the optimization process.

Including this term in our \TGV model we can penalize high depth discontinuities at homogeneous regions and allow sharp depth edges at corresponding texture differences.
A similar combination of \TGV and weighting was used by Ranftl \etal. \cite{Ranftl2012} for passive stereo reconstruction.
With the additional edge tensor information the optimization result leads to sharper and more defined edges in our solution. Further, the regions where the depth data is interpolated are filled out more reasonably. 

The final energy is defined as a combination of data term (\ref{equ:data_term}) and the \TGV term (\ref{equ:energy_tgv_2}) with anisotropic diffusion (\ref{equ:anisoTensor}):
\begin{align}
\begin{split}
  \min_{u,v} \left\lbrace \alpha_{1} \int_{\Omega_H} \mathrm |T^{\frac{1}{2}} (\nabla u - v)| \,\mathrm dx + \alpha_{0} \int_{\Omega_H} \mathrm |\nabla v| \,\mathrm dx + \right. \\
  \left. \int_{\Omega_H} \mathrm w|(u - \Ds)|^{2} \,\mathrm dx \right\rbrace .
\end{split}
  \label{equ:energy_final_single_tgv}
\end{align}

\subsection{Primal-Dual Optimization} \label{ssec:optimization}
The proposed optimization problem (\ref{equ:energy_final_single_tgv}) is convex but non smooth due to \TGV regularization term and the zeros in the weighting operator $w$.
To find a fast, global optimal solution for our problem we use the primal-dual energy minimization scheme, as proposed in \cite{Chambolle2011, Esser2010}. We reformulate the non-smooth problem in a convex-concave saddle-point problem applying the \LF. The optimization problem can be efficiently minimized through gradient descent. The transformed saddle-point problem of our energy functional (\ref{equ:energy_final_single_tgv}) is given by
\begin{align}
\begin{split}
  &\min_{u\in \mathbb{R}^{MN},v\in \mathbb{R}^{2MN}} \max_{p \in P, q \in Q} \alpha_1 \langle T^{\frac{1}{2}} (\nabla u - v), p \rangle  +  \\ 
& \qquad \quad \alpha_0 \langle \nabla v , q \rangle + \sum_{i,j \in \Omega} w_{i,j}( u_{i,j} - D_{S{i,j}} )^{2},
\end{split}
  \label{equ:energy_pd}
\end{align}
introducing the dual variables $p$ and $q$. The feasible sets of these variables are defined by
\begin{align}
  &P = \left\lbrace p \colon \Omega_{H} \rightarrow \mathbb{R}^{2} | \enskip \| p_{i,j} \| \leq 1, \enskip \forall i,j \in \Omega_H \right\rbrace, \\
  &Q = \left\lbrace q \colon \Omega_{H} \rightarrow \mathbb{R}^{4} | \enskip \| q_{i,j} \| \leq 1, \enskip \forall i,j \in \Omega_H \right\rbrace.
\end{align}
This formulation is used in the primal-dual algorithm, where the primal and dual variables are iteratively optimized for the individual pixels in three steps. First, the dual variables $p$ and $q$ are updated 
using gradient ascend. Second, the primal variables are updated using gradient-descent. Third, the primal variables are refined in an overrelaxation step. 
The step sizes are chosen s.t. $u^{0} = \Ds$, $v^{0},p^{0},q^{0} = 0$, $\sigma_p > 0$, $\sigma_{q} > 0$, $\tau_u > 0$ and $\tau_v > 0$. For any iteration $n \geq 0$ the steps are calculated according to
\begin{align}
  \begin{cases}
    p^{n+1} = \mathcal{P}_{p} \left\lbrace p^{n} + \sigma_p \alpha_{1} \left( T^{1/2} (\nabla \bar{u}^{n} - \bar{v}^{n}) \right) \right\rbrace \\ 
    q^{n+1} = \mathcal{P}_{q} \left\lbrace q^{n} + \sigma_q \alpha_{0} \nabla \bar{v}^{n} \right\rbrace \\
    u^{n+1} = \dfrac {u^{n} + \tau_u\left( \alpha_1 \nabla^{\mathrm T} T^{1/2} p^{n+1} + w \Ds \right)}{1 + \tau_u w} \\
    v^{n+1} = v^{n} + \tau_v \left(\alpha_0 \nabla^{\mathrm T} q^{n+1} + \alpha_1 T^{1/2} p^{n+1} \right) \\
    \bar{u}^{n+1} = u^{n+1} + \theta( u^{n+1} - \bar{u}^{n} ) \\
    \bar{v}^{n+1} = v^{n+1} + \theta( v^{n+1} - \bar{v}^{n} )
  \end{cases}
  \label{equ:pd_fusion_algo}
\end{align}
until a stopping criterion is reached. To fulfill the convex optimality condition in the dual update step, the projection operators $\mathcal{P}_{p}$ and $\mathcal{P}_{q}$ for $p$ and $q$ are calculated through
\begin{align}
\begin{split}
  \mathcal{P}_p \left\lbrace \tilde{p}_{i,j} \right\rbrace = \frac{\tilde{p}_{i,j}}
  {\max\left(1, \left| \tilde{p}_{i,j} \right| \right)}, \\
  \mathcal{P}_q \left\lbrace \tilde{q}_{i,j} \right\rbrace = \frac{\tilde{q}_{i,j}}
  {\max\left(1, \left| \tilde{q}_{i,j} \right| \right)}.
      \end{split}
  \label{equ:resolvent_p}
\end{align}

In practice the relaxation parameter $\theta$ is updated in every iteration, according to \cite{Chambolle2011}, and the optimal step sizes are calculated using preconditioning, as proposed in \cite{Pock2011}. Therewith, we achieve a fast and guaranteed convergence to the global optimal solution for different tensor conditions. The gradient and divergence operators are approximated using forward/backward differences with Neumann and Dirichlet boundary conditions, respectively.
\savenotes
\begin{table*}

  \begin{small}
  \begin{center}
    \begin{tabular}{@{}l*{13}{c}@{}}
      \toprule
      \multirow{2}{*}{  } & \multicolumn{4}{c}{\textit{Art}} & \multicolumn{4}{c}{\textit{Books}} & \multicolumn{4}{c}{\textit{Moebius}} & \multirow{2}{*}{Avg.Time [s]}\\ \cmidrule(lr){2-5}\cmidrule(lr){6-9}\cmidrule(lr){10-13}
       & x2 & x4 & x8 & x16 & x2 & x4 & x8 & x16 & x2 & x4 & x8 & x16 & \\ \midrule
      Nearest & 4.65 & 5.01 & 5.71 &7.10 & 4.30 & 4.68 & 4.85 & 5.23 & 5.08 & 5.20 & 5.31 & 5.65 & - \\
      Bilinear & 3.09 & 3.59 & 4.39 & 5.91 & 2.91 & 3.12 & 3.34 & 3.71 & 3.21 & 3.45 & 3.62 & 4.00 & -\\
      Yang \etal \cite{Yang2007} & 1.36 & 1.93 & \uline{2.45} & 4.52 & 1.12 & 1.47 & \uline{1.81} & \uline{2.92} & 1.25 & 1.63 & \uline{2.06} & 3.21 & -\\
      He \etal \cite{He2010} & 1.92 & 2.40 & 3.32 & 5.08 & 1.60 & 1.82 & 2.31 & 3.06 & 1.77 & 2.03 & 2.60 & 3.34 & 23.89\\
      Diebel and Thrun \cite{Diebel2006} & 1.62 & 2.24 & 3.85 & 5.70 & 1.34 & 2.08 & 2.85 & 3.54 & 1.47 & 2.29 & 3.09 & 3.81 & -\\
      Chan \etal \cite{Chan2008} & 1.83 & 2.90 & 4.75 & 7.70 & 1.04 & \uline{1.36} & 1.94 & 3.07 & 1.17 & 1.55 & 2.28 & 3.55 & 3.02\footnote{This is an extrapolation of the runtime the authors report on images of size $800\times 600$.}\\ 
      Park \etal \cite{Park2011} & \uline{1.24} & \uline{1.82} & 2.78 & \uline{4.17} & \uline{0.99} & 1.43 & 1.98 & 3.04 & \uline{1.03} & \uline{1.49} & 2.13 & \uline{3.09} & 24.05 \\ \midrule
      \textit{OURS} & \textbf{0.84} & \textbf{1.29} & \textbf{2.06} & \textbf{3.56} & \textbf{0.51} & \textbf{0.75} & \textbf{1.16} & \textbf{1.89} & \textbf{0.57} & \textbf{0.90} & \textbf{1.38} & \textbf{2.15} & \textbf{1.94} \\ \bottomrule 
    \end{tabular}
  \end{center}
  \end{small}
  \caption{Quantitative comparison on the Middlebury 2007 datasets with added noise. The error is measured as \textit{RMSE} of the pixel disparity for four different magnification factors ($\times2$, $\times4$, $\times8$, $\times16$). The best result for each dataset and upscaling factor is highlighted and the second best is underlined.}
\label{tab:exp_middlebury}
\end{table*}
\begin{figure*}
  \begin{center}
  \hspace*{-4pt}
  \subfloat[RGB]{
    \includegraphics*[width=0.141\textwidth]{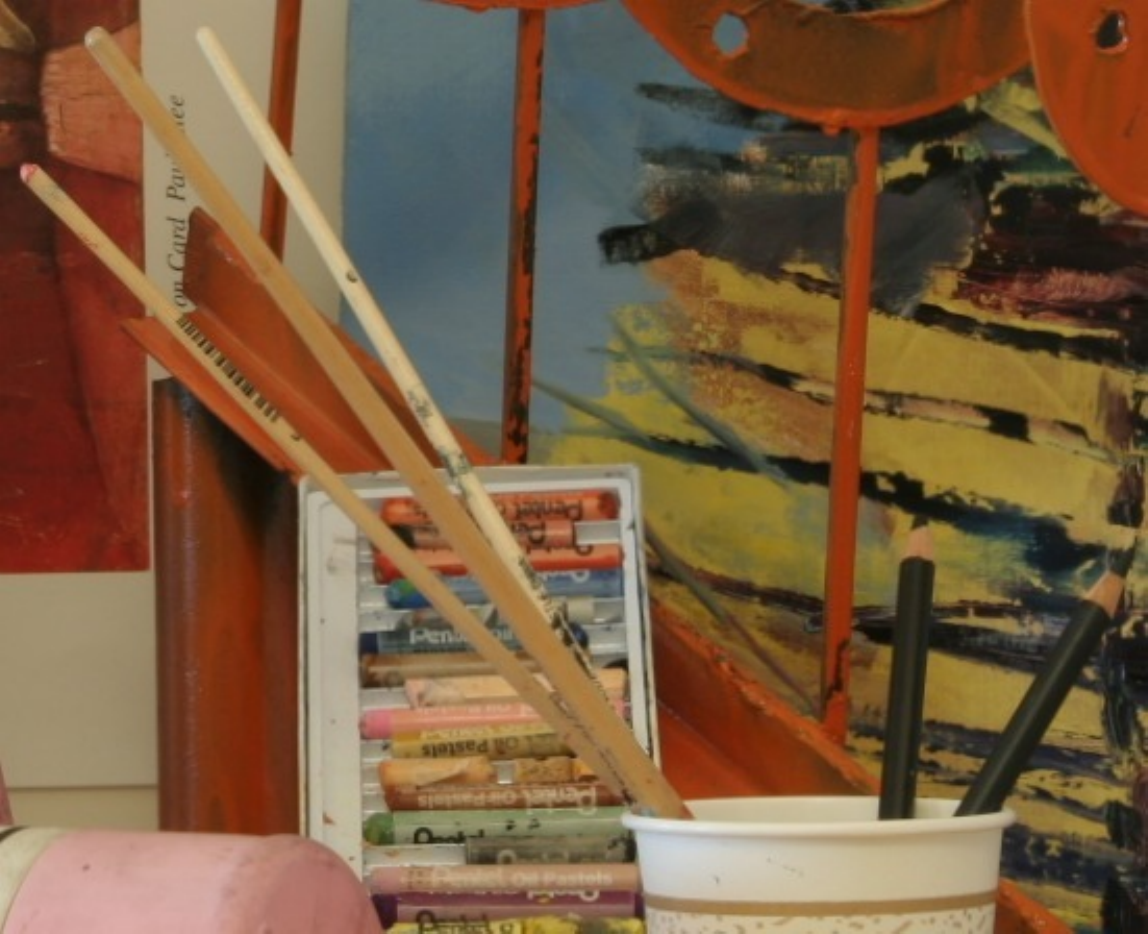}
  } \hspace*{-6pt}
  \subfloat[input depth]{
    \includegraphics*[width=0.141\textwidth]{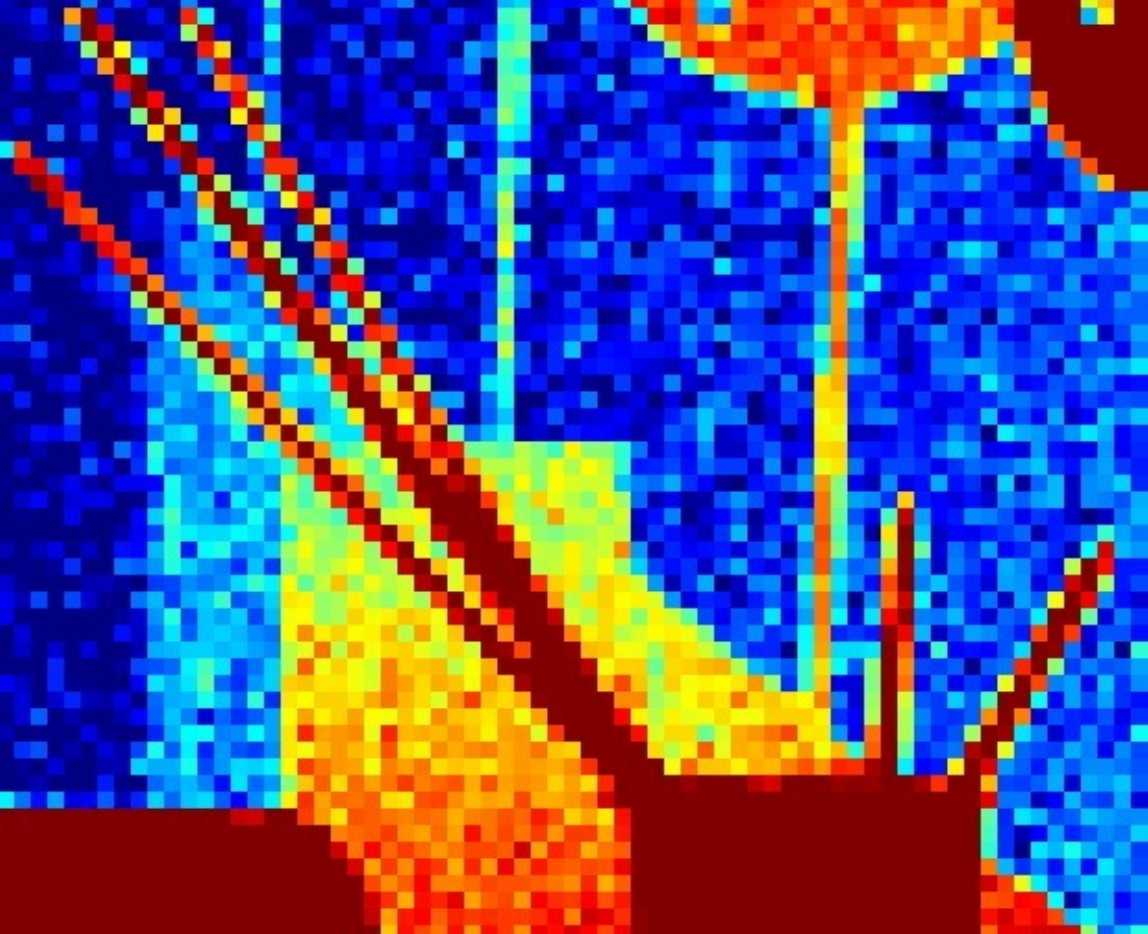}
  } \hspace*{-6pt}
  \subfloat[Diebel]{
    \includegraphics*[width=0.141\textwidth]{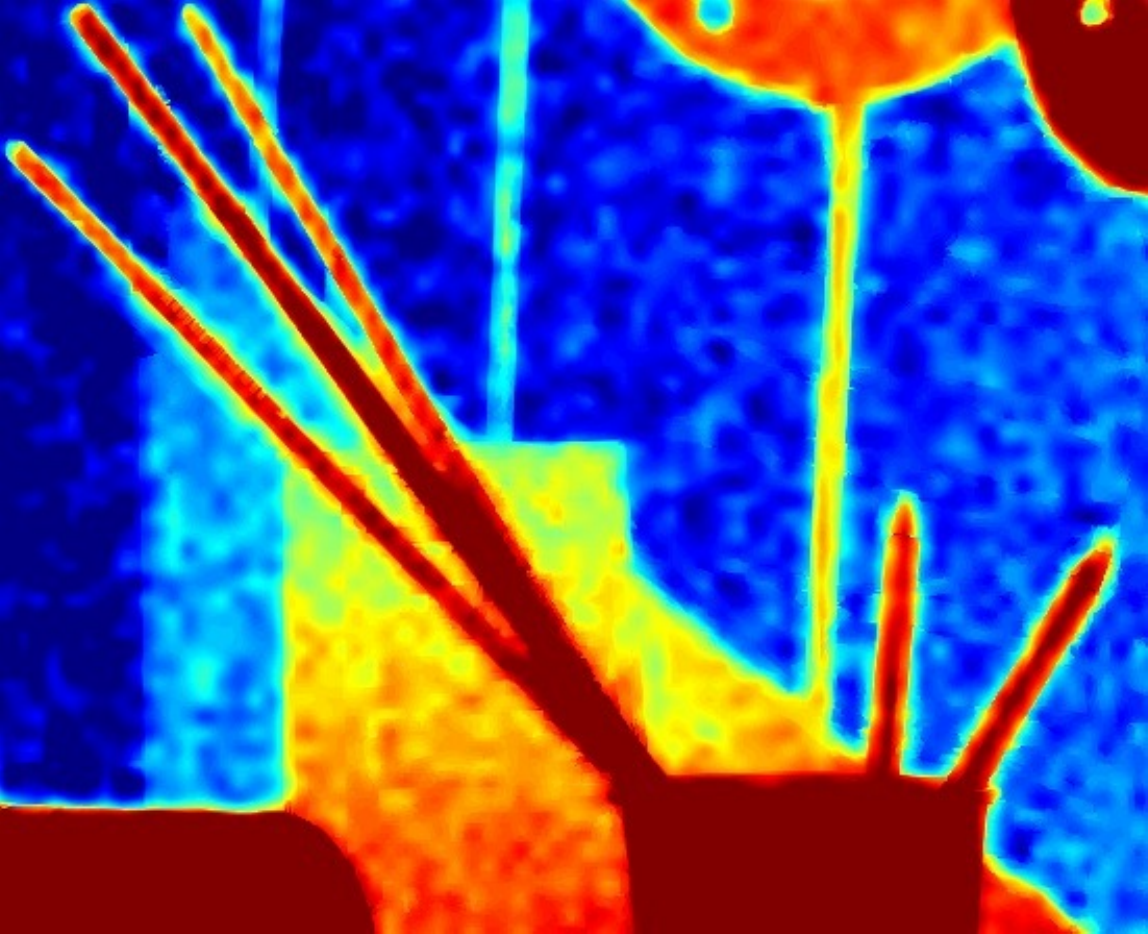}
  } \hspace*{-6pt}
  \subfloat[Chan]{
    \includegraphics*[width=0.141\textwidth]{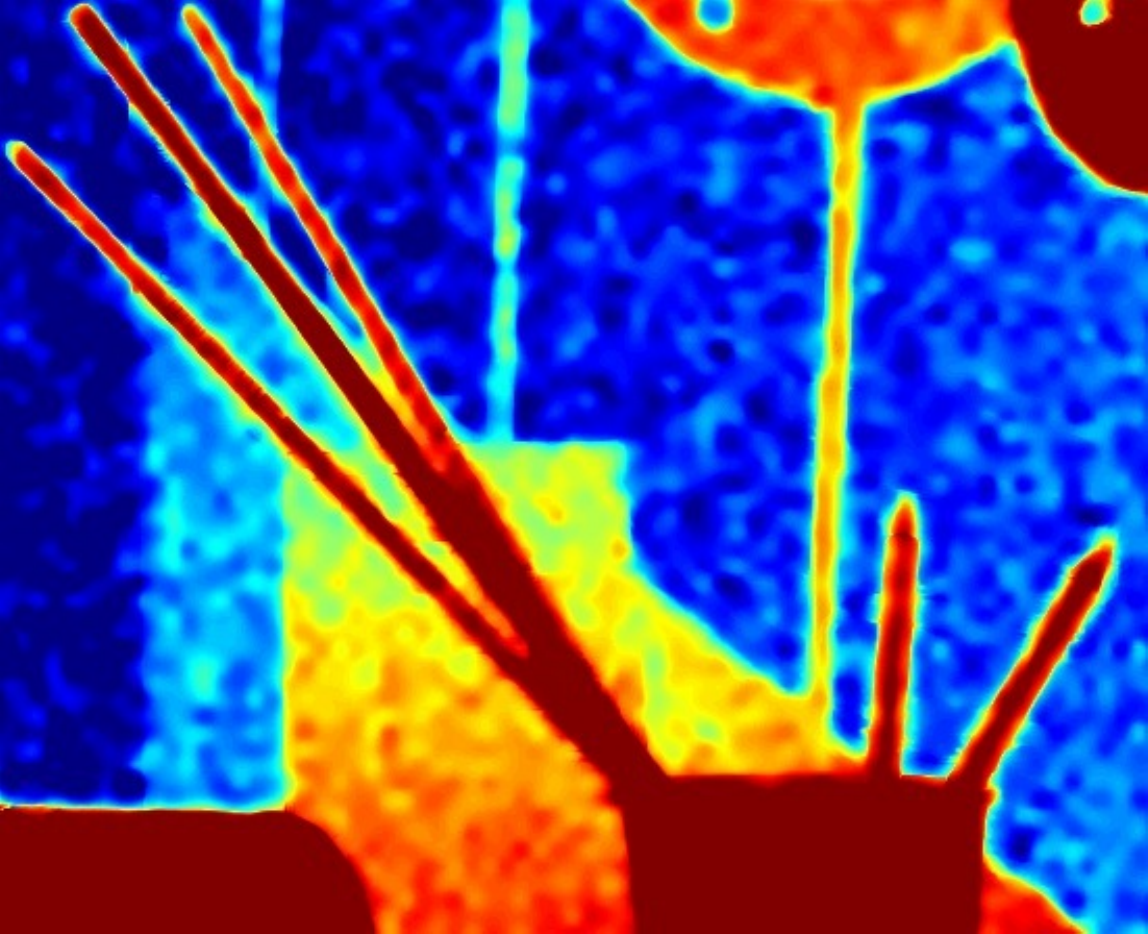}
  } \hspace*{-6pt}
  \subfloat[Park]{
    \includegraphics*[width=0.141\textwidth]{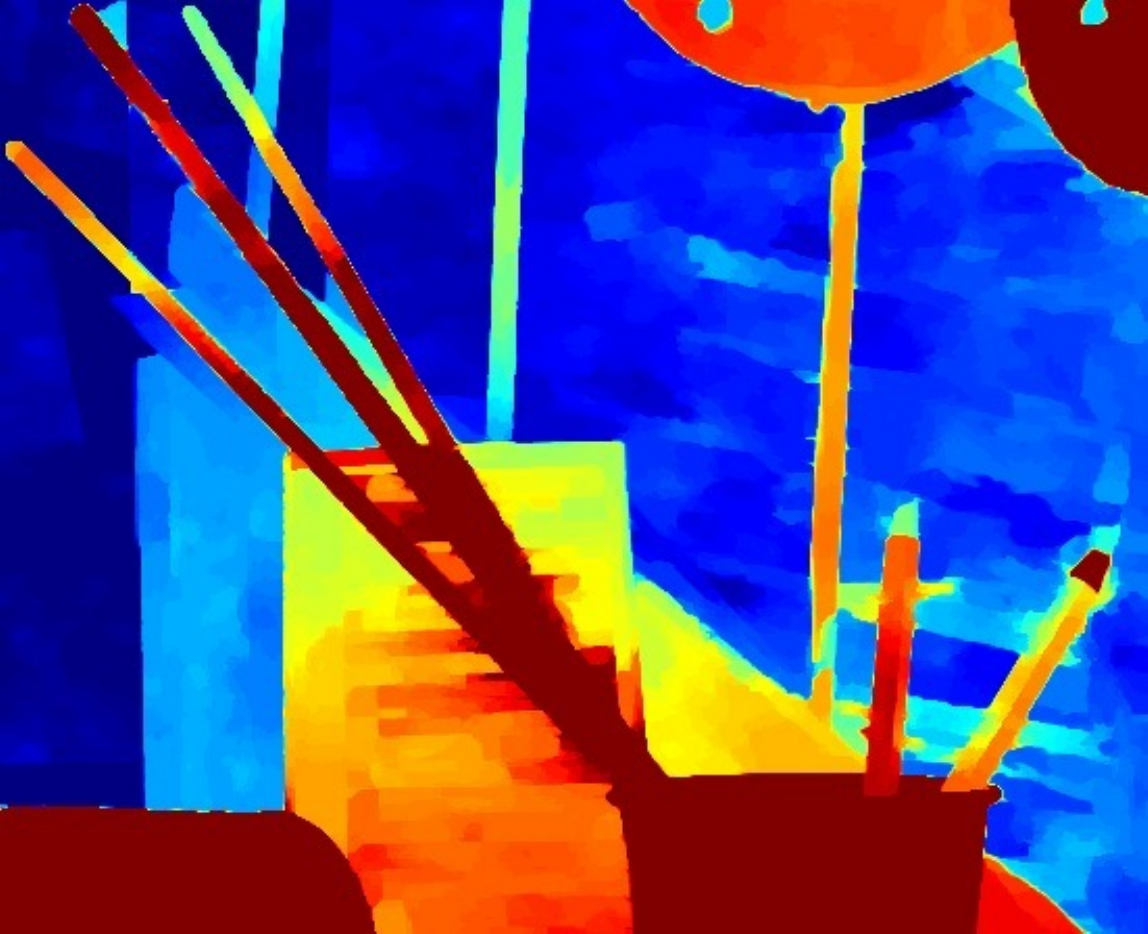}
  } \hspace*{-6pt}
  \subfloat[\textit{OURS}]{
    \includegraphics*[width=0.141\textwidth]{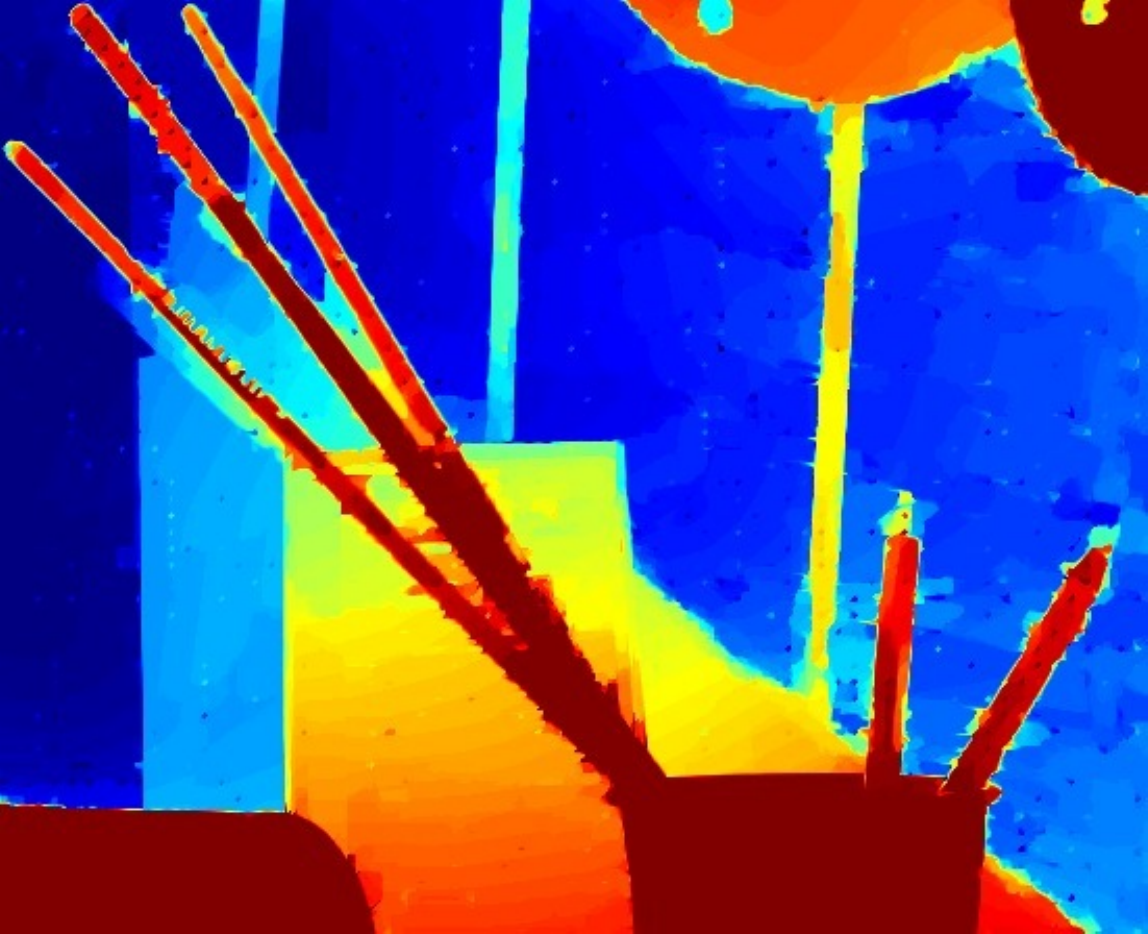}
  } \hspace*{-6pt}
  \subfloat[Groundtruth]{
    \includegraphics*[width=0.141\textwidth]{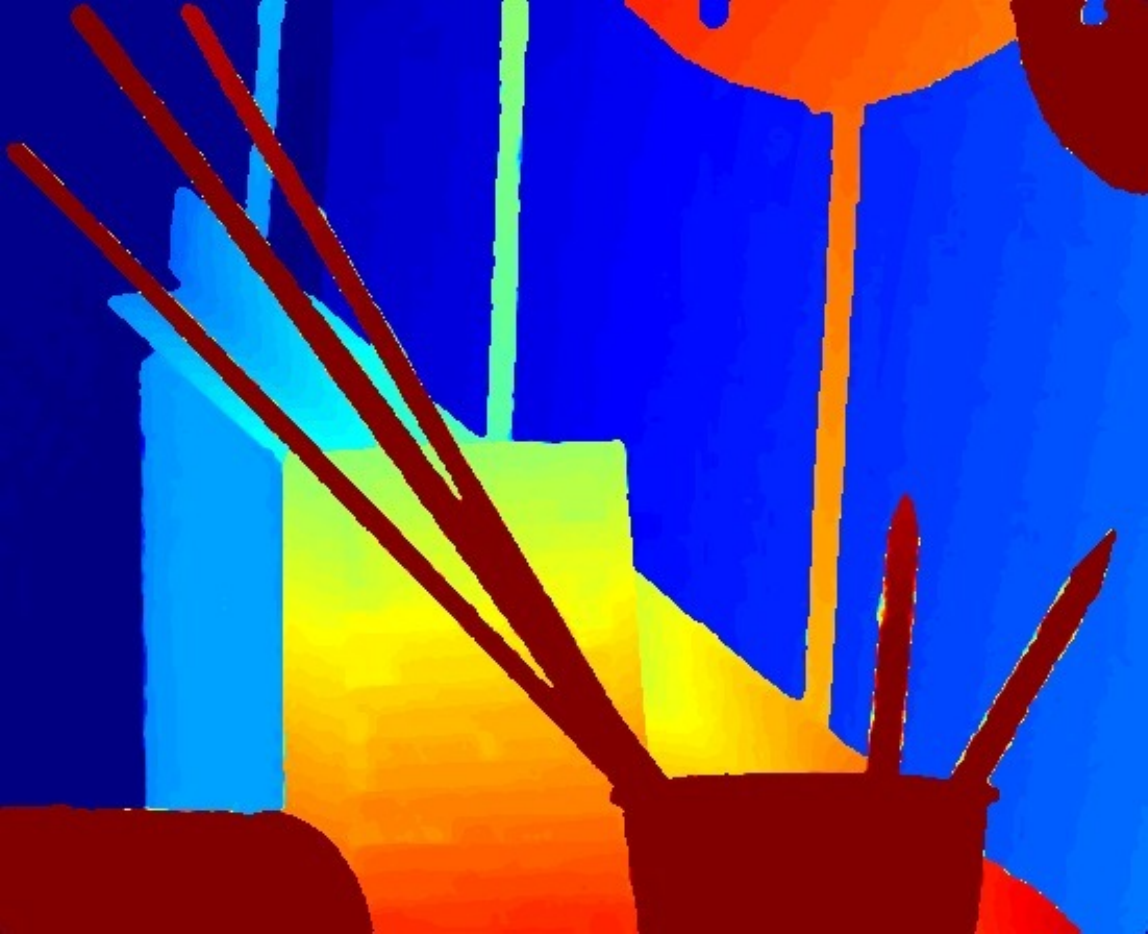}
  }
\end{center}
  \caption{Visual comparison of $\times8$ \SR on a snippet of the Middlebury \textit{Art} dataset including fine structures. (a) RGB intensity image, (b) low resolution input image (enlarged using nearest neighbor upsampling). (c) \uSR using MRF proposed by Diebel and Thrun \cite{Diebel2006}. (d) Adaptive bilateral upsampling proposed by Chan \etal \cite{Chan2008}. (e) Nonlocal means upsampling proposed by Park \etal \cite{Park2011}. (f) Our \SR method using image guided anisotropic TGV. The results in (c) and (d) still suffer from noise. (e) removes noise but suffers from edge bleeding especially at small structure boundaries. Our method removes noise and preserves sharp object edges.
  }
\label{fig:exp_middlebury}
\end{figure*}
\ResetAbbrevs{All}
\section{Evaluation}
In this section, we show a quantitative and qualitative evaluation of our \SR method. 
For an extensive evaluation we investigate the performance compared to \SOTA approaches on the simulated Middlebury 2007 datasets \cite{Hirschmuller2007,Scharstein2007} in terms of speed and accuracy.
Beyond this simulations, we evaluate our method on real data with highly accurate groundtruth measurements. In our experiments we use a $2 \times 2$ gradient operator to calculate the intensity image gradients. The tensor parameters $\beta$ and $\gamma$ as well as the \textit{TGV} parameters $\alpha_0$ and $\alpha_1$ are manually set once for each upsampling factor and are constant in synthetic and the real world evaluations.
 
\spewnotes
\subsection{Middlebury Benchmark Evaluation}
An exhaustive evaluation of our method in terms of quantitative and qualitative comparison is made using input images from the Middlebury datasets \cite{Hirschmuller2007,Scharstein2007}. We use the disparity image as groundtruth and the original RGB intensity image as input for our anisotropic diffusion tensor. Park \etal \cite{Park2011} provides low resolution input depth images with different downsampling factors ($\times2$, $\times4$, $\times8$, $\times16$). To simulate the acquisition process, these input images contain additional Gaussian noise with a standard deviation that increases with the disparity. Using these datasets we are able to compare our results with the \MRF based approach of Diebel and Thrun \cite{Diebel2006}, the bilateral filtering with cost volume refinement of Yang \etal \cite{Yang2007}, the guided image filtering approach of He \etal \cite{He2010}, the noise-aware bilateral filter approach by Chan \etal \cite{Chan2008} and the non-local means filtering by Park \etal \cite{Park2011}.
Further, we compare the results to common interpolation methods.
The confidence measure $w$ in our functional is set to $1$ for all depth points. The parameters $\alpha_0$ and $\alpha_1$ have been kept fixed for all datasets and have been empirically chosen for $\times2$ / $\times4$ / $\times8$ / $\times16$ as $0.154$, $0.023$ / $0.05$, $0.0056$ / $0.267$, $0.03$ / $0.267$, $0.03$. 


This experiment gives an objective comparison on the robustness, accuracy and speed of a variety of different algorithms. The numerical results for this experiment in terms of the \RMSE and computation time are shown in Table \ref{tab:exp_middlebury}. A visual comparison for the different methods is given in Figure \ref{fig:exp_middlebury}. Further quantitative comparisons to other depth \SR methods on the Middlebury 2003 and 2007 datasets can be found in the supplemental material. 

\paragraph*{Discussion}
What can be clearly seen is that our method delivers an upsampling quality that is superior compared to \SOTA methods at a lower computation time. This quality improvement originates from a \TGV regularization combined with an anisotropic diffusion tensor. The higher order regularization better captures the surface of real world scenes, while the anisotropic diffusion tensor delivers a more defined guidance of the high resolution intensity data compared to a simple scalar weighting.

While the Middlebury datasets are popular to evaluate depth upsampling methods, they neglect some important properties of real acquisition setups. Typically, depth and intensity data do not originate from the same sensor and are therefore not aligned. Further, real low resolution depth sensors measure depth data with a more complex acquisition noise which can not be simulated by adding simple Gaussian noise. Therefore, we create a novel benchmarking dataset based on real sensor data.
\subsection{Benchmarking based on Real Sensor Data}\label{ssec:benchmark}
The evaluation on real acquisitions is made using different scenes acquired with a \TOF and an intensity camera simultaneously. For depth measurements we use a PMD Nano \TOF camera delivering a $120 \times 160$ dense depth and IR amplitude image \cite{PMDCBN}. The intensity image is acquired by a CMOS camera with a sensor size of $810 \times 610$ pixel. 

\paragraph*{Camera Calibration}
Calibration of the intensity camera and the \TOF camera is a crucial part in our \SR system since the quality of the calibration directly affects the accuracy of the \SR result. 
The intrinsic and extrinsic camera parameters are calibrated similar to the method of Zhang \cite{Zhang2000}. A planar target with circular feature points and known geometry is acquired from different viewpoints. Through the known correspondence between the feature points on the target plane and in the image, focal length, principal point and radial and tangential distortions are estimated. To calibrate the intrinsic parameters of the \TOF camera, the low resolution IR amplitude image $\Il\colon \Omega_{\mathrm L} \subseteq \mathbb{R}^{2}$ is used. 
The rotation and translation between intensity and \TOF camera is estimated by establishing a geometric correspondence through the feature points on the planar target.

\begin{figure}[ht]
      \newlength{\figwidth}
      \newlength{\figheight}
      \setlength{\figwidth}{0.82\columnwidth} 
      \setlength{\figheight}{0.6\columnwidth}
      \newcommand{\lw}{0.5pt}
      \newcommand{\lwd}{1.25pt}
	\input{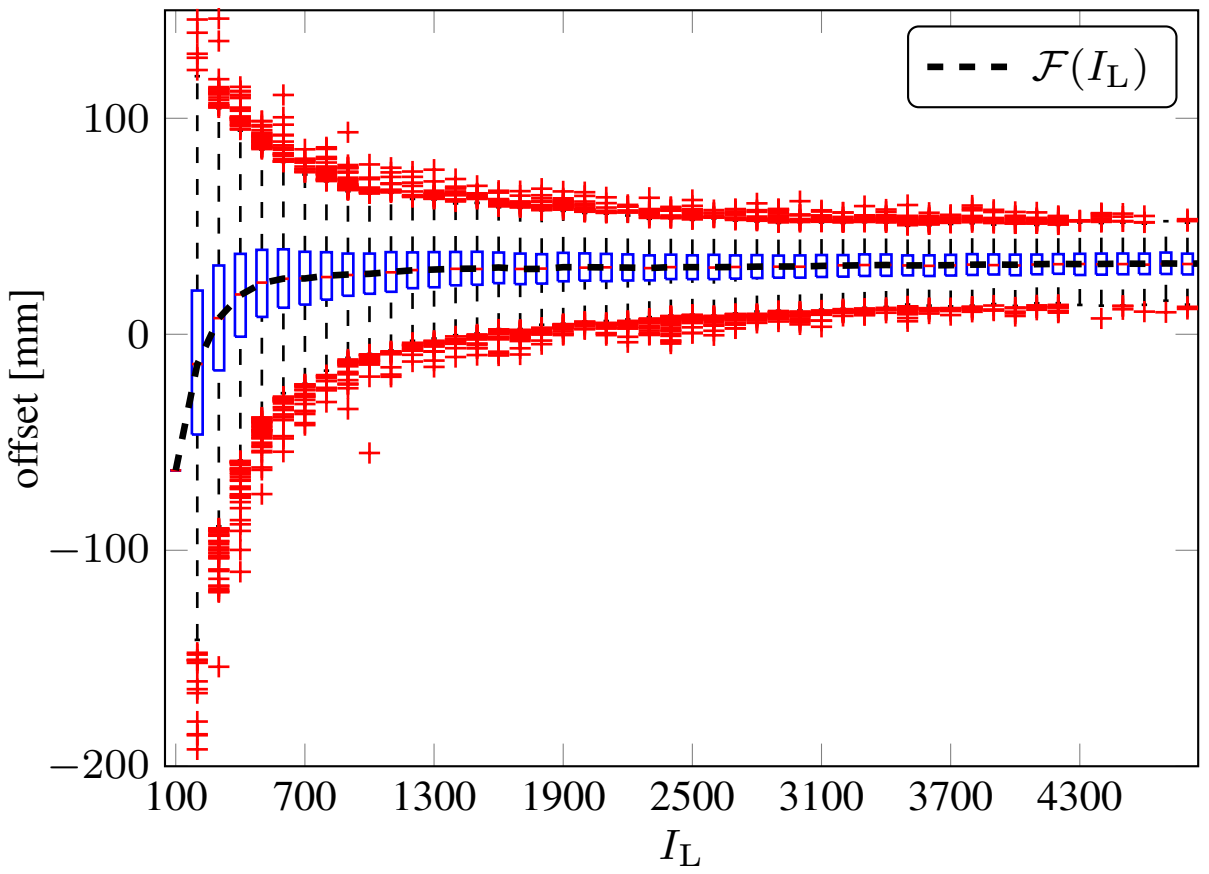}
  \caption{Correlation between measured IR amplitude and depth measurement error of TOF acquisitions. The standard deviation of the measurements increases with decreasing IR intensities.}
\label{fig:amplitued2offset}
\end{figure}

\begin{figure}[ht]
  \begin{center}
      \includegraphics[width=\columnwidth]{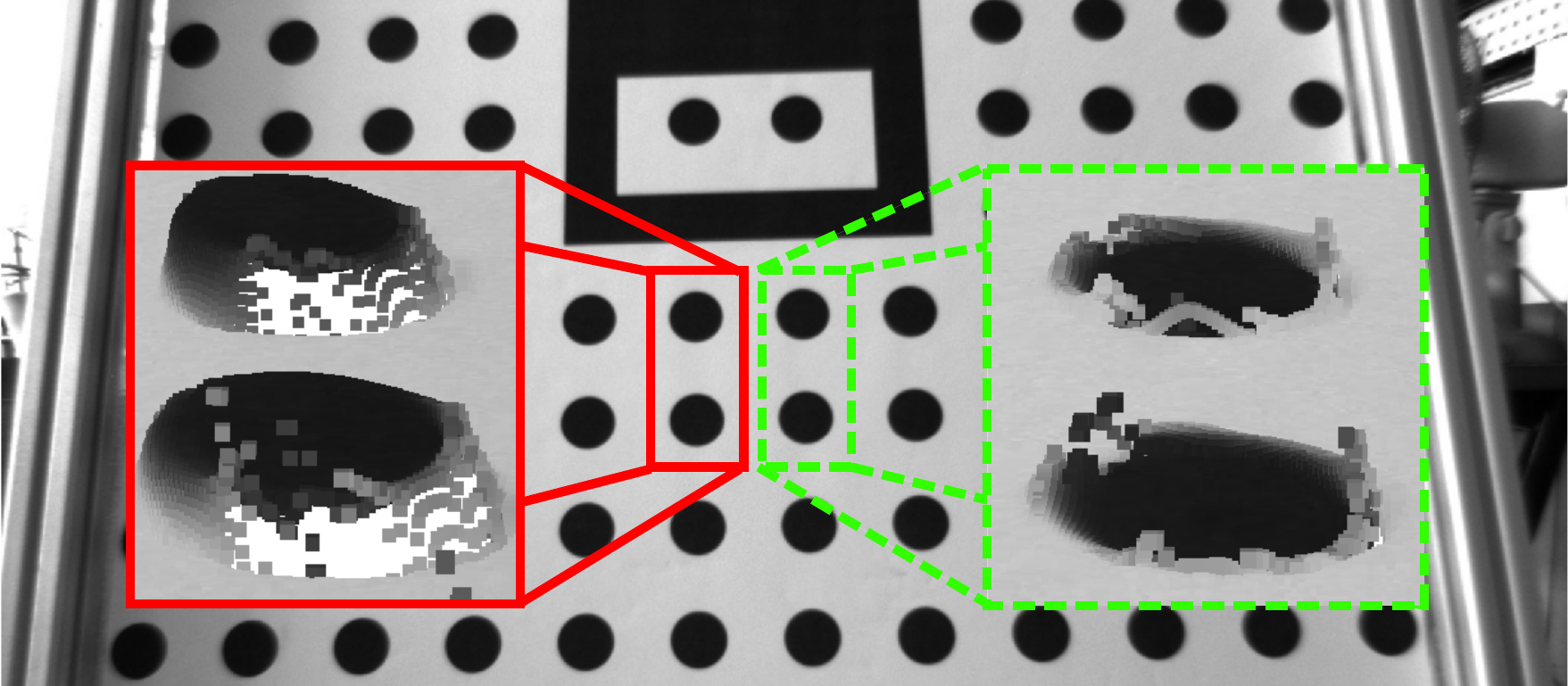}
  \end{center}
  \caption{Compensation for amplitude based depth error of TOF cameras. Depth estimates on dark surfaces are underestimated (see red/full box). Using our depth calibration, we can compensate for that error (see green/dashed box).}
  \label{fig:compensation}
\end{figure}

In addition to the intrinsic and extrinsic camera parameters, both the \TOF depth is calibrated and a depth confidence value is calculated through the 3D projections of the planar feature points. Because \TOF cameras measure depth through active illumination, the depth measurement certainty increases with the measured amplitude \cite{Fuchs2008}. 
Through a comparison of the very accurate 3D measurements of the calibration points and the measured \TOF depth points a dependence between the acquired IR amplitude image and the measurement error can be established, as shown in Figure~\ref{fig:amplitued2offset}.
With this information we can estimate an amplitude-dependent 3D measurement offset value $\Delta d_{i,j} = \mathcal{F}(\mat I_{\mathrm{L}\, i,j}) , \mathcal{F}: \mathbb{R}\rightarrow \mathbb{R}$ and a confidence measure $w$ which decreases with increasing noise. Depth estimates where the amplitude is very low are underestimated. This effect can be clearly seen on our black and white calibration target (see Figure~\ref{fig:compensation}). After compensation $\tilde d_{i,j}  = d_{i,j} + \Delta d_{i,j}$, the difference between bright and dark surface reconstructions can be minimized.

We assume a linear correspondence between the IR amplitude and the depth error below a full confidence threshold $I_{max}$, as shown in Figure \ref{fig:amplitued2offset}. Hence, the depth confidence weighting is calculated through the \TOF amplitude image \Il by
$ w_{i,j} = 
\begin{cases}
I_{L{i,j}}/I_{max} \quad &\text{if} \enskip I_{L\,i,j} < I_{max} \\
I_{L{i,j}} = 1 \quad &\text{else}
\end{cases} \quad \forall i,j \in \Omega_L$. The full confidence value is empirically set to $I_{max}=1000$ ($16 \unit{bit}$ amplitude image \Il). 

\paragraph*{Novel Real-World Dataset}
The groundtruth measurements are generated using a structured light scanner which consists of two $2048 \times 2048$ pixel high-speed intensity cameras and one high-speed projector. The depth uncertainty at the given baseline is $1.2 \unit{mm}$. To get a dense depth map, multiple acquisitions with slightly displaced projection angles are fused together.
The acquired scenes are chosen to incorporate structures with high texture variations (see \emph{Books} scene) as well as thin wiry elements (\emph{Shark} and \emph{Devil} scenes) to evaluate the upsampling accuracy. All scenes lie in the depth range of $0.8 - 1.2 \unit{m}$ which reflects the operation distance of modern \TOF cameras.

\begin{figure*}
  \newlength\imglen
  \setlength\imglen{2.96cm}
\begin{center}
\includegraphics[height=\imglen]{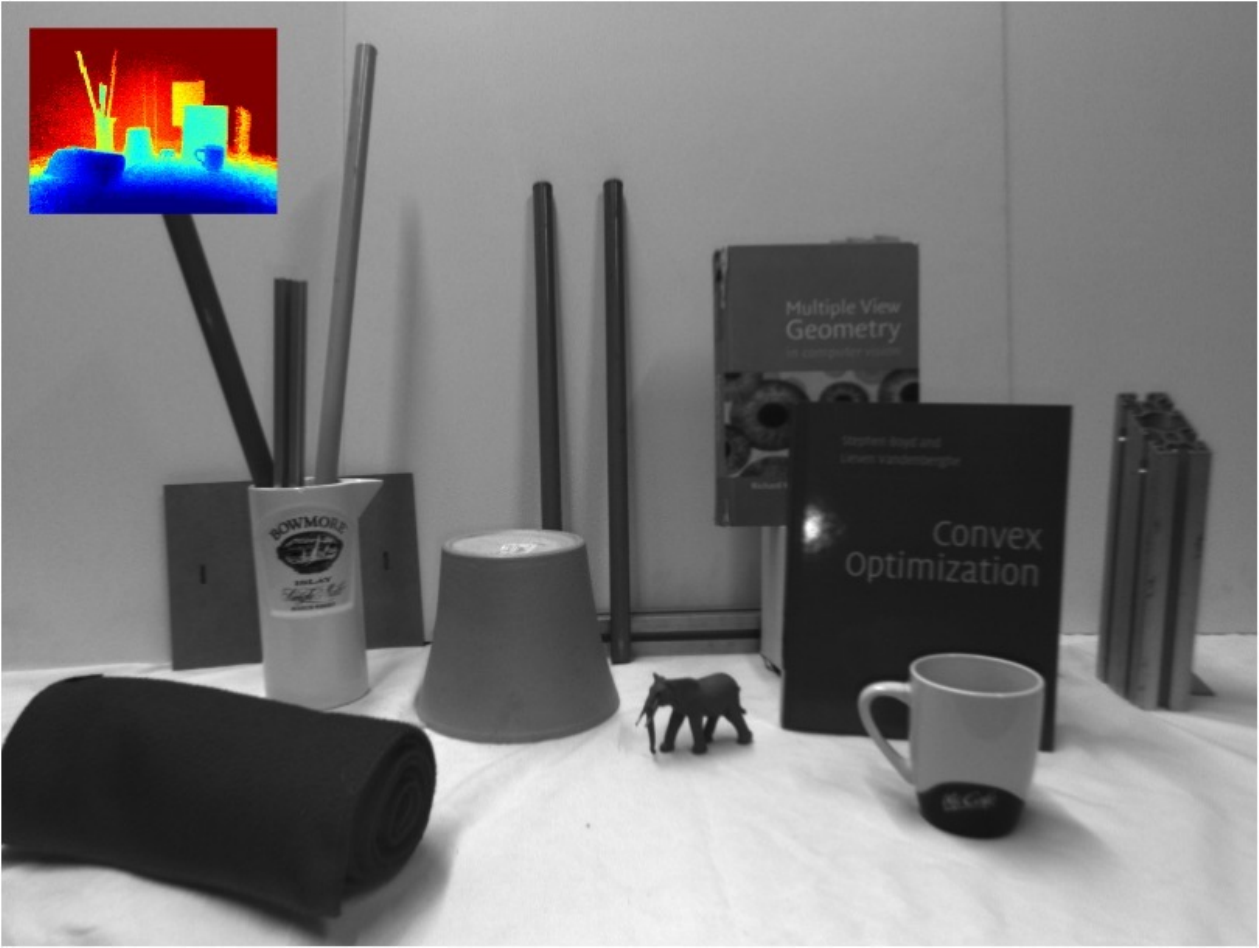}
\includegraphics[height=\imglen]{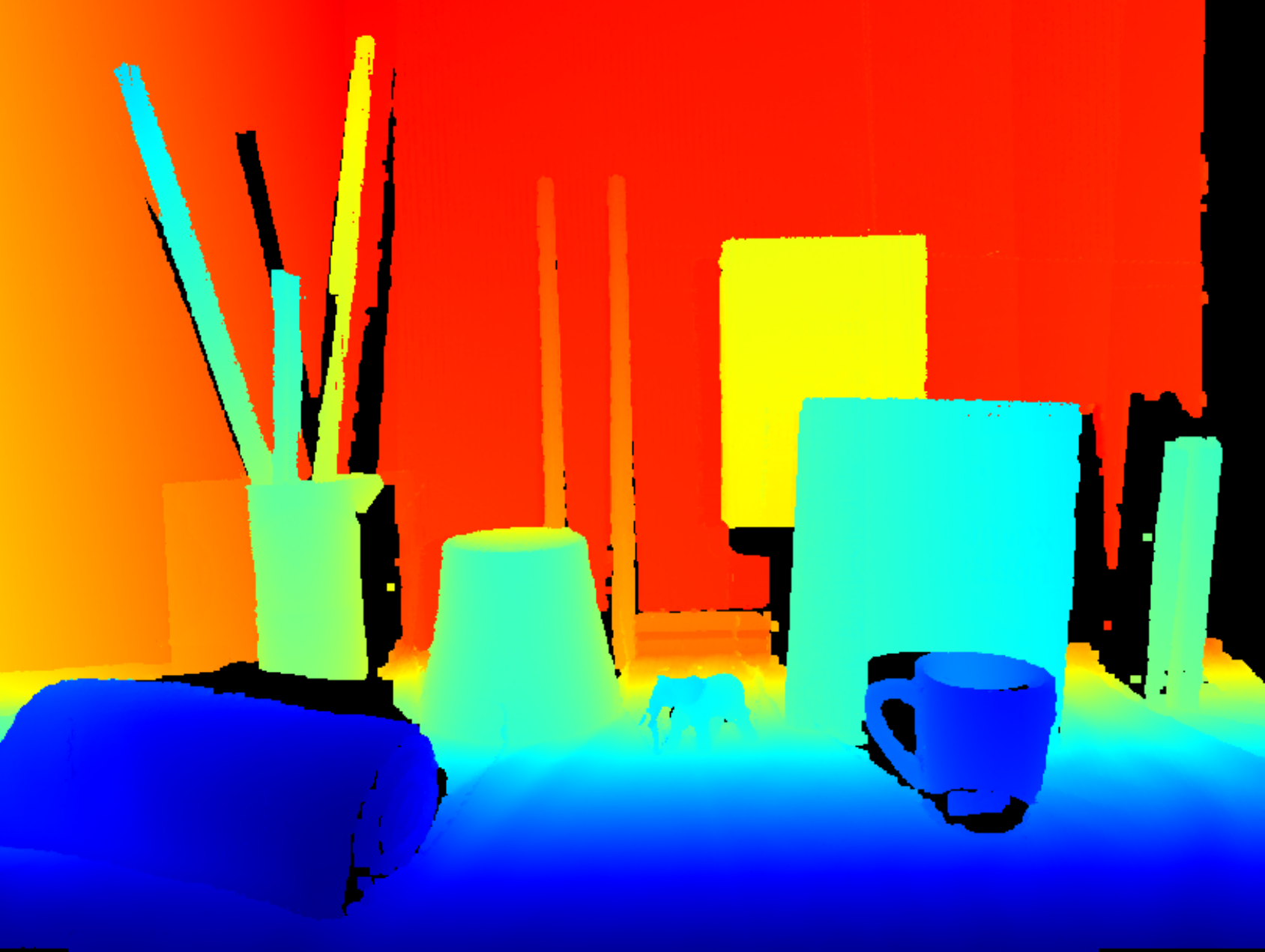}
\includegraphics[height=\imglen]{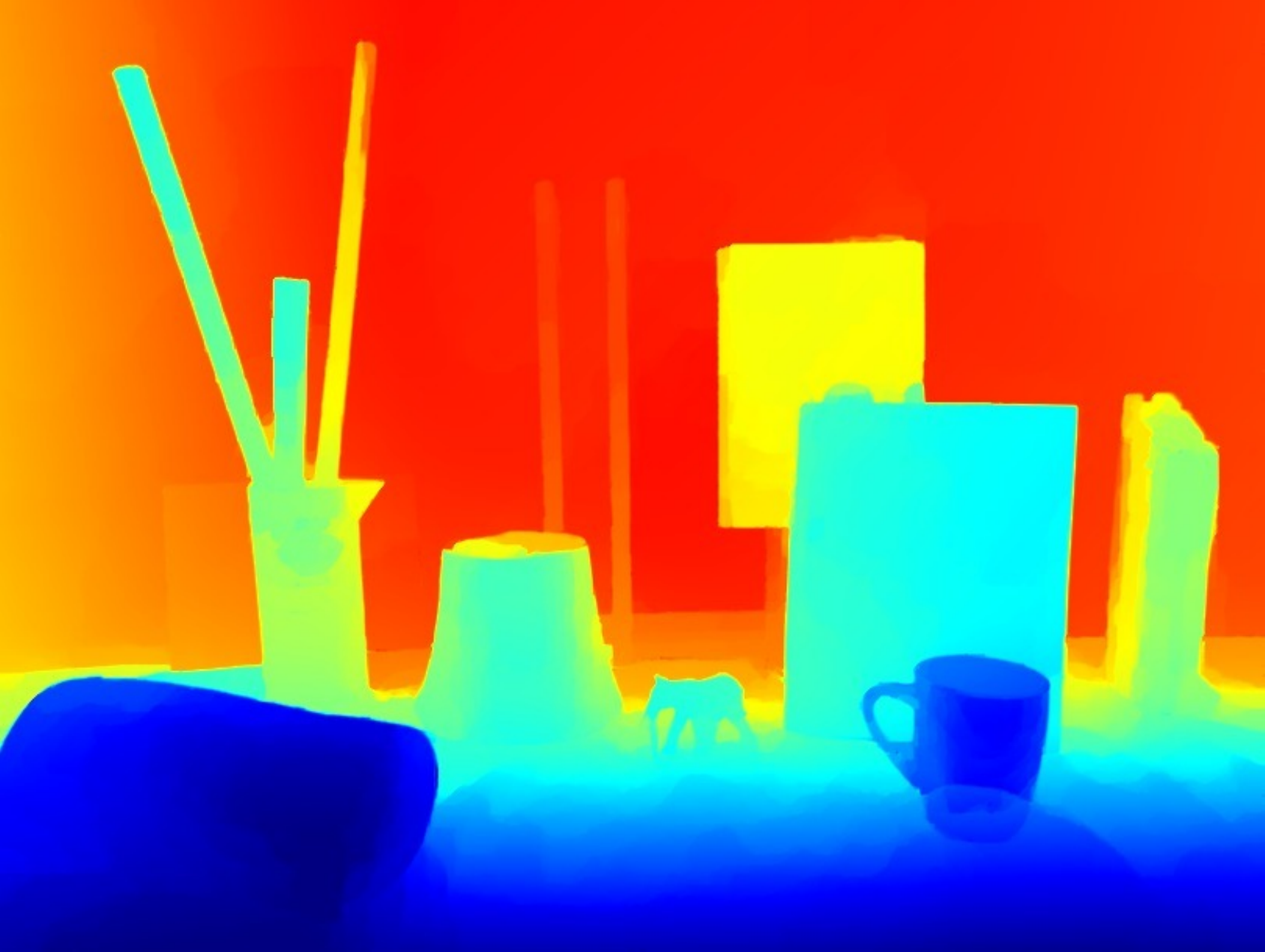}
\includegraphics[height=\imglen]{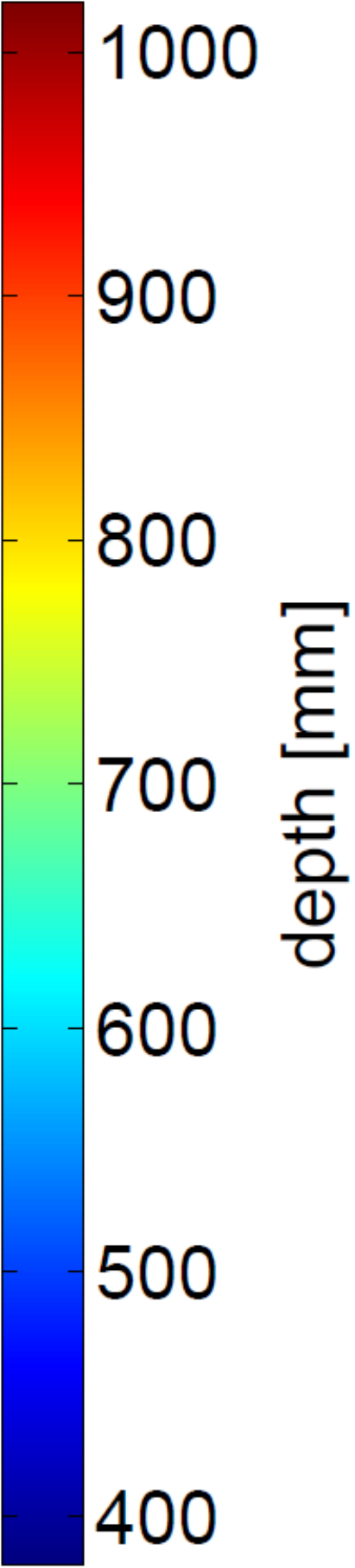}
\includegraphics[height=\imglen]{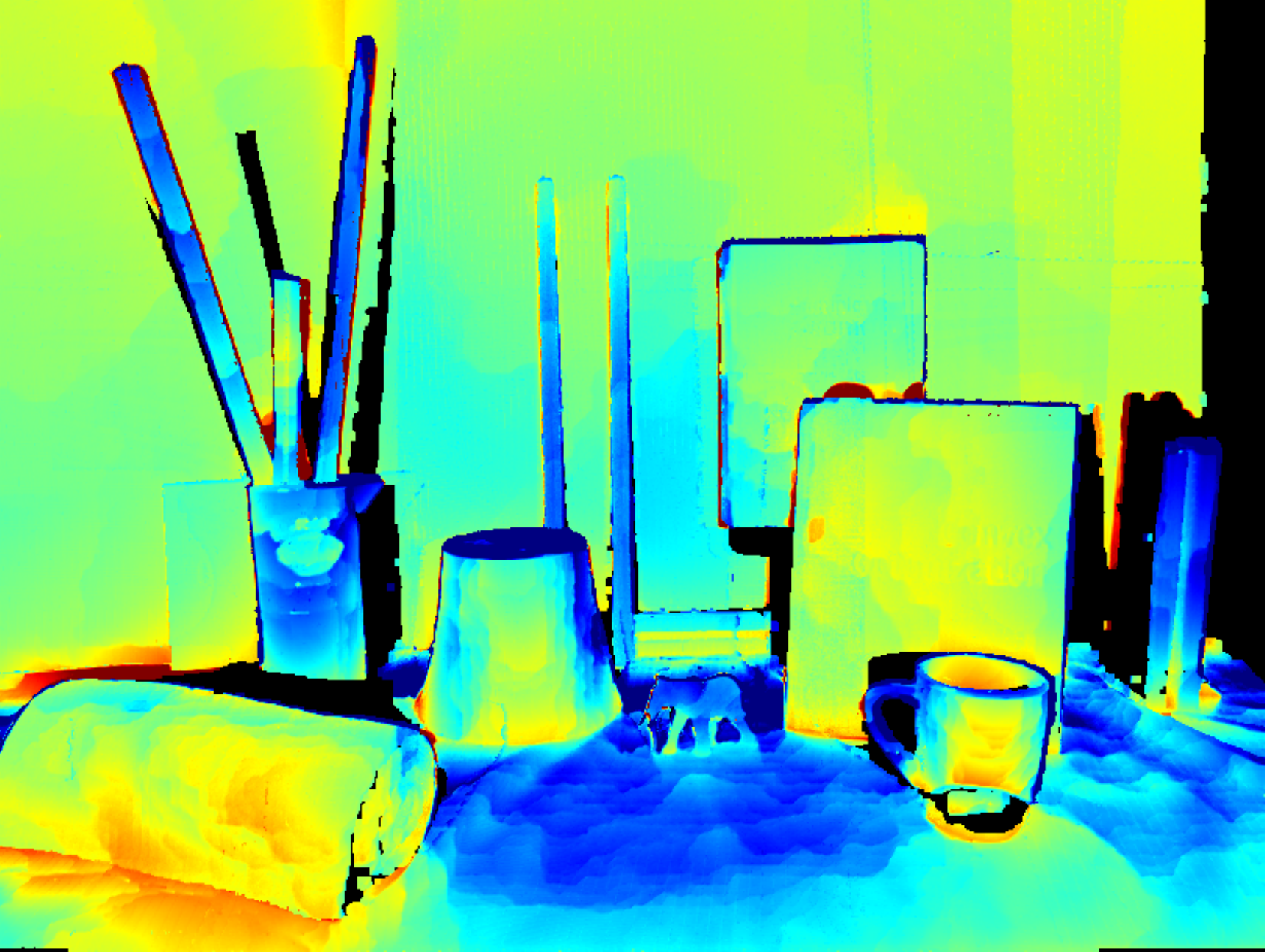}
\includegraphics[height=\imglen]{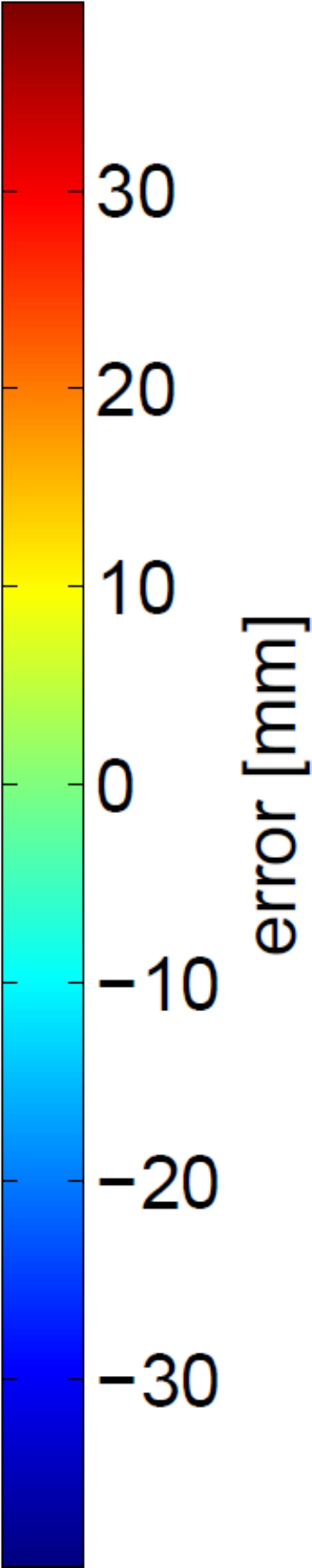}\\
\includegraphics[height=\imglen]{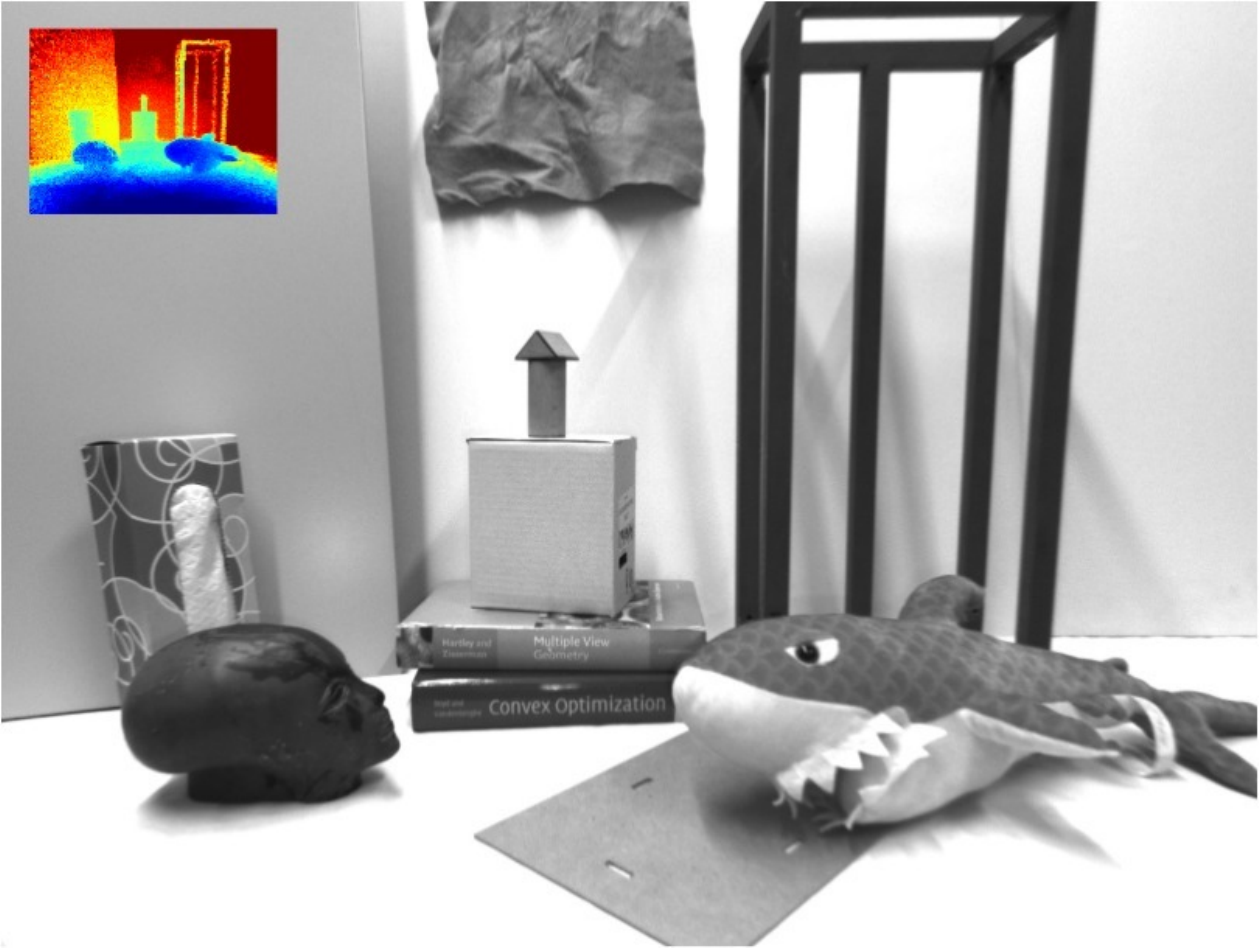}
\includegraphics[height=\imglen]{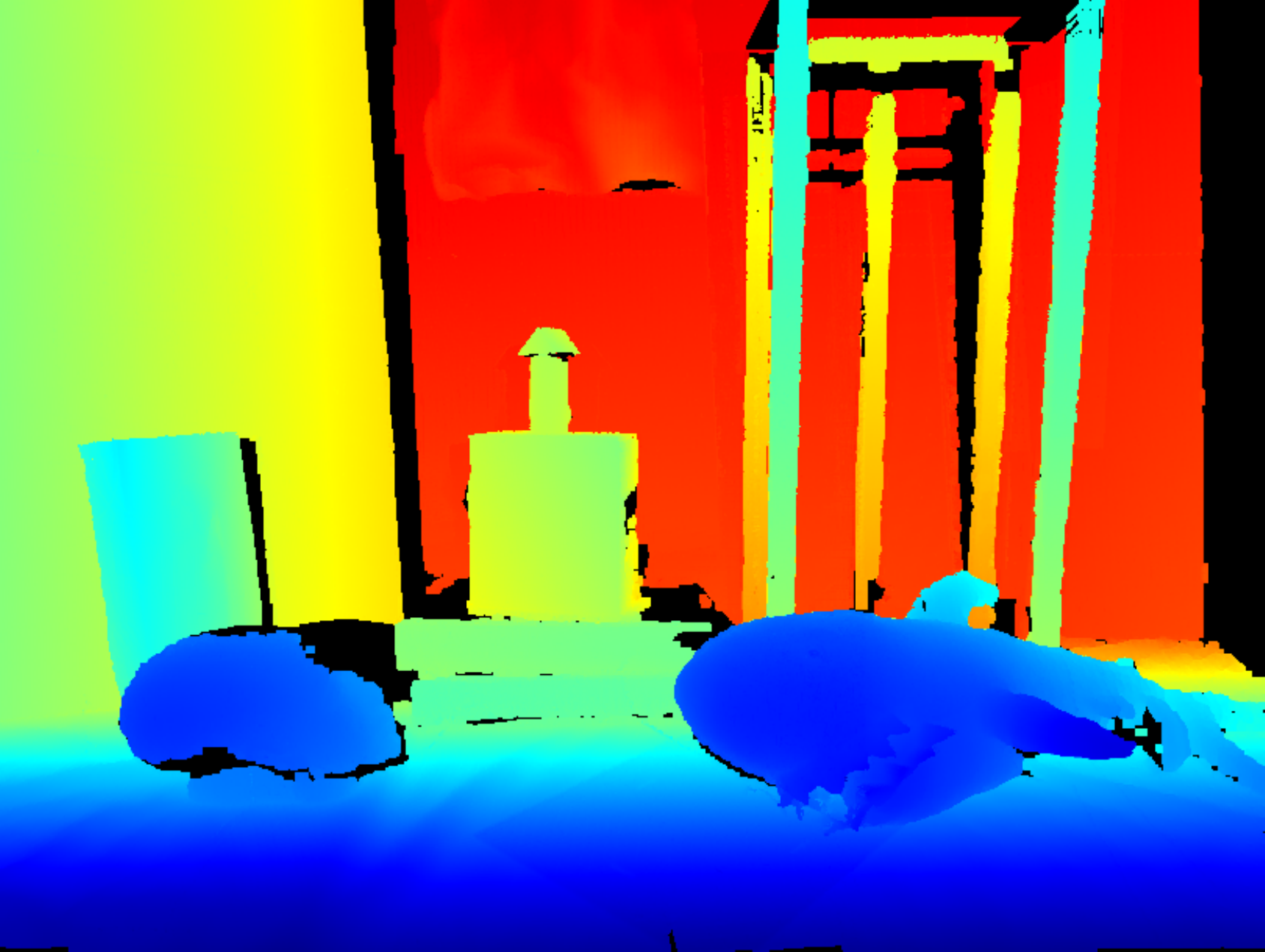}
\includegraphics[height=\imglen]{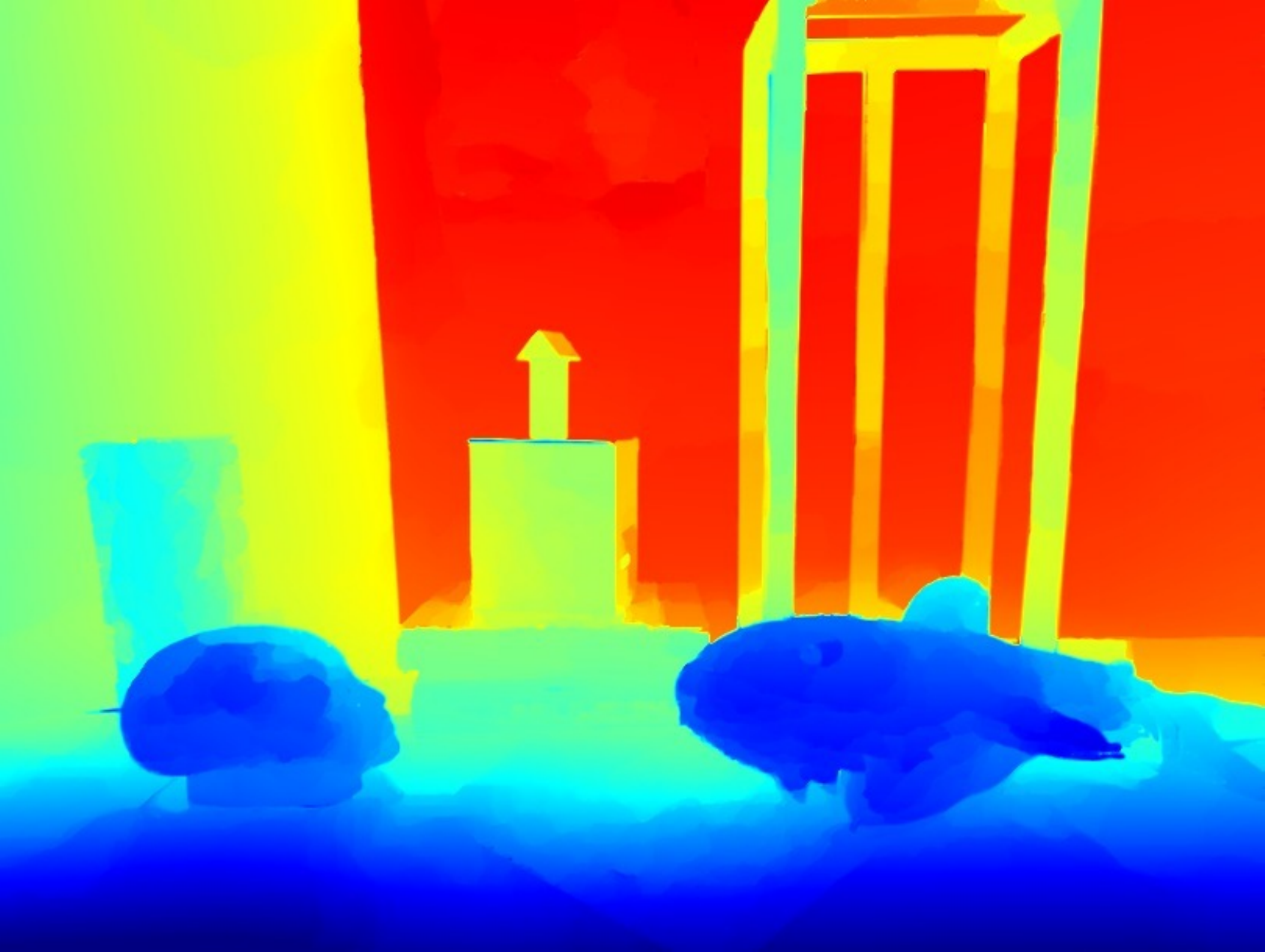}
\includegraphics[height=\imglen]{colorbar_depth_with}
\includegraphics[height=\imglen]{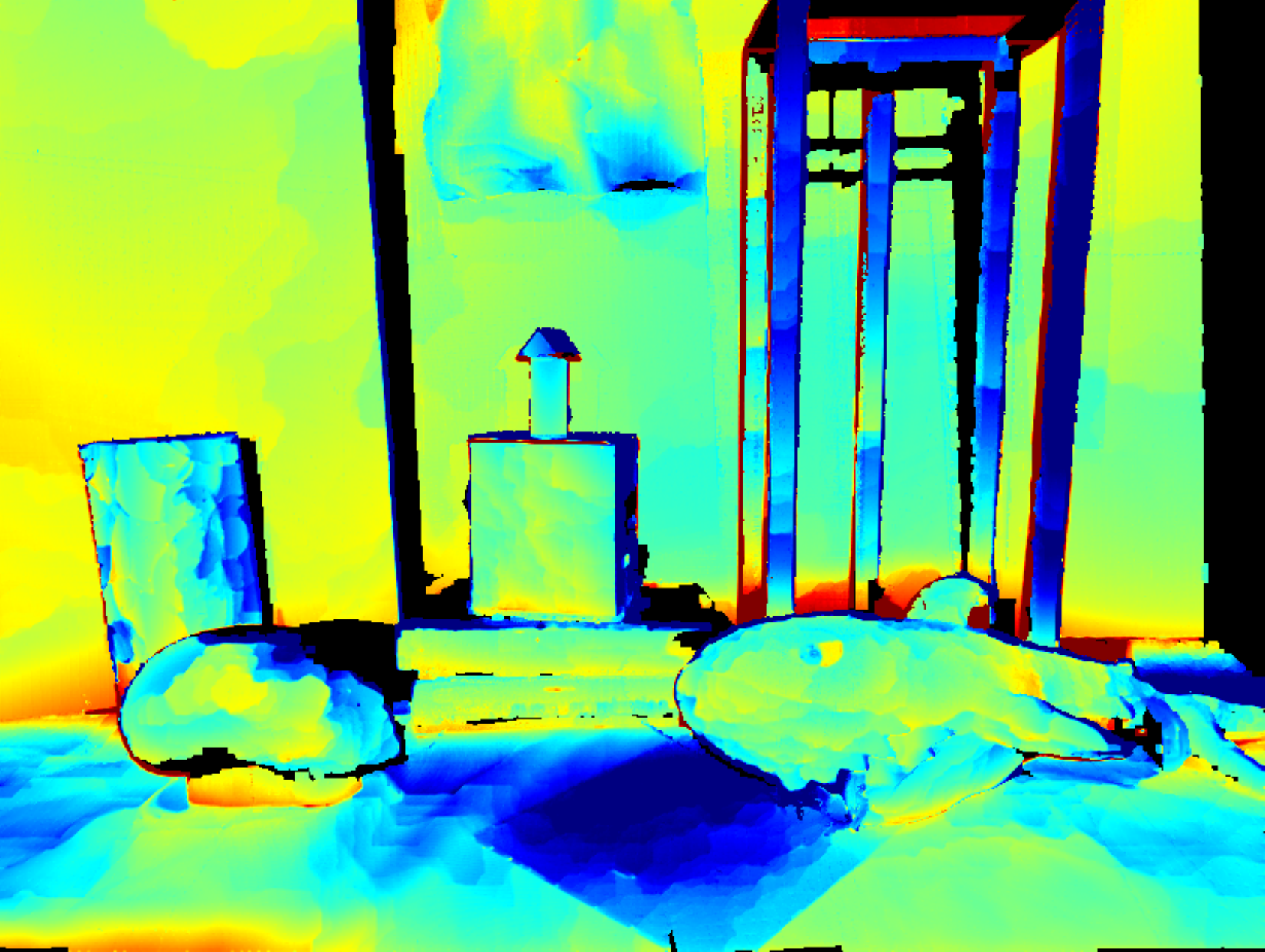}
\includegraphics[height=\imglen]{colorbar_error}\\\vspace*{-3.5mm}\hspace*{-2.2 pt}
\subfloat[Intensity + \textit{ToF}]{\includegraphics[height=\imglen]{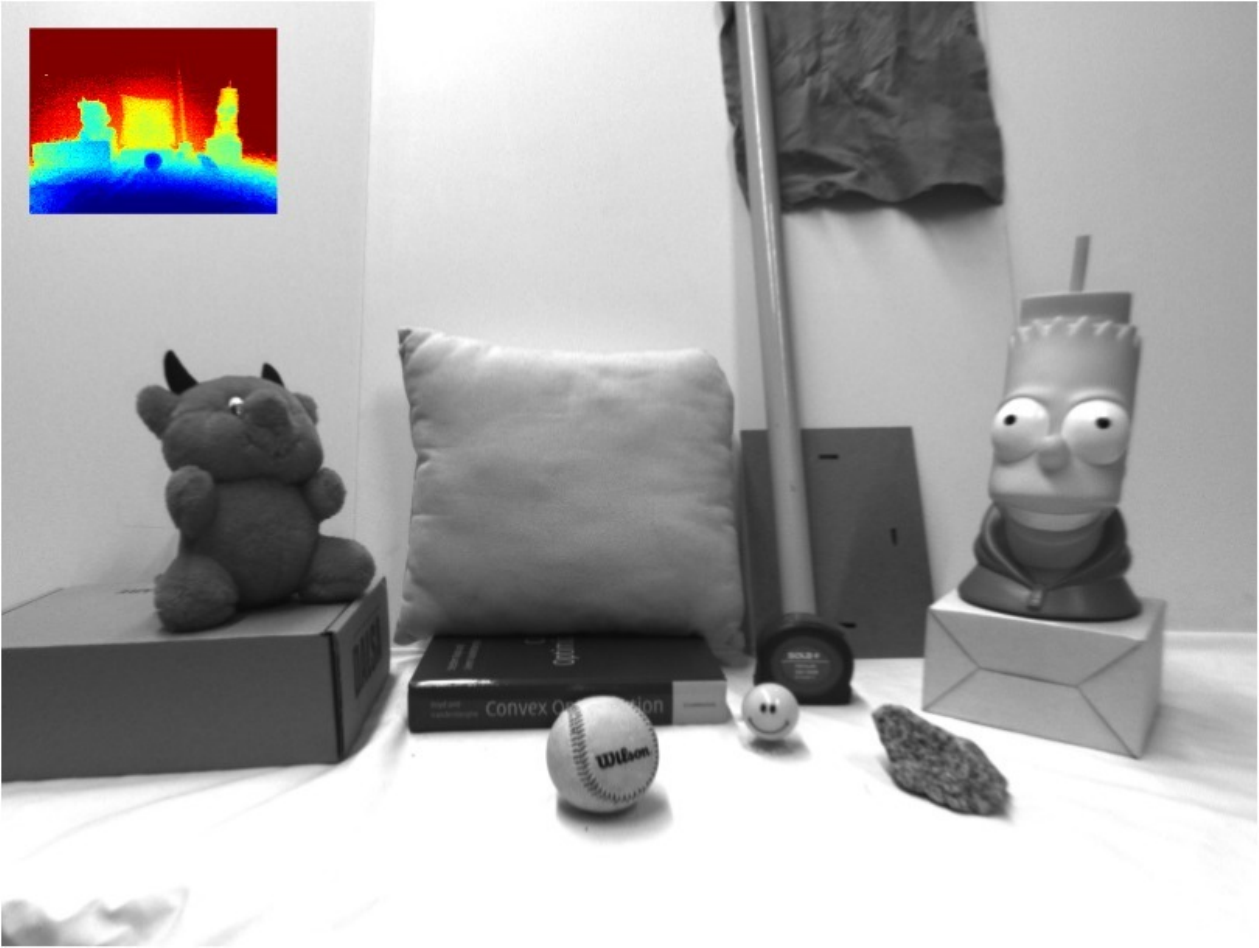}}\hspace*{0.1pt}
\subfloat[Groundtruth]{\includegraphics[height=\imglen]{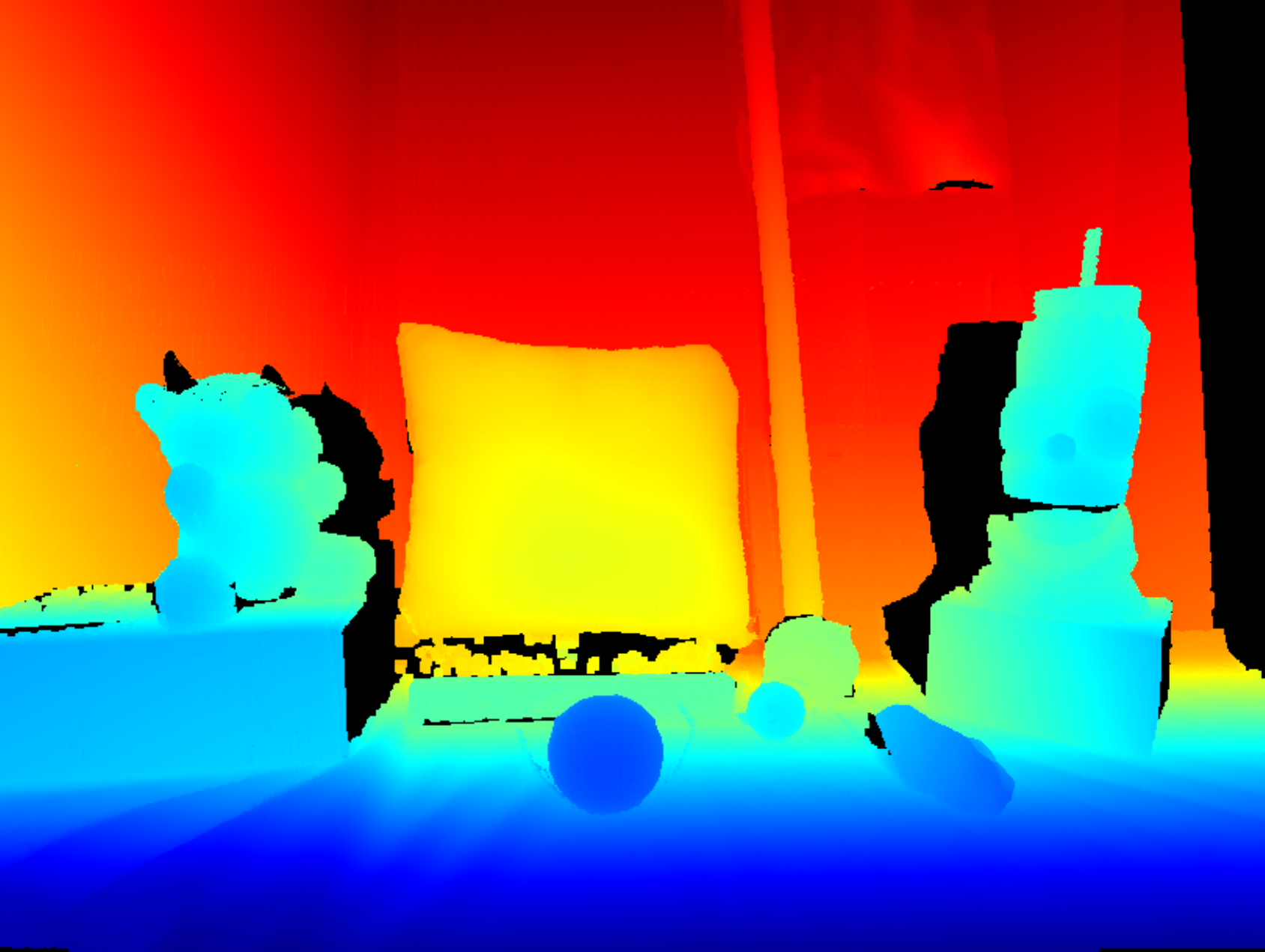}}\hspace*{0.1pt}
\subfloat[\textit{OURS}]{\includegraphics[height=\imglen]{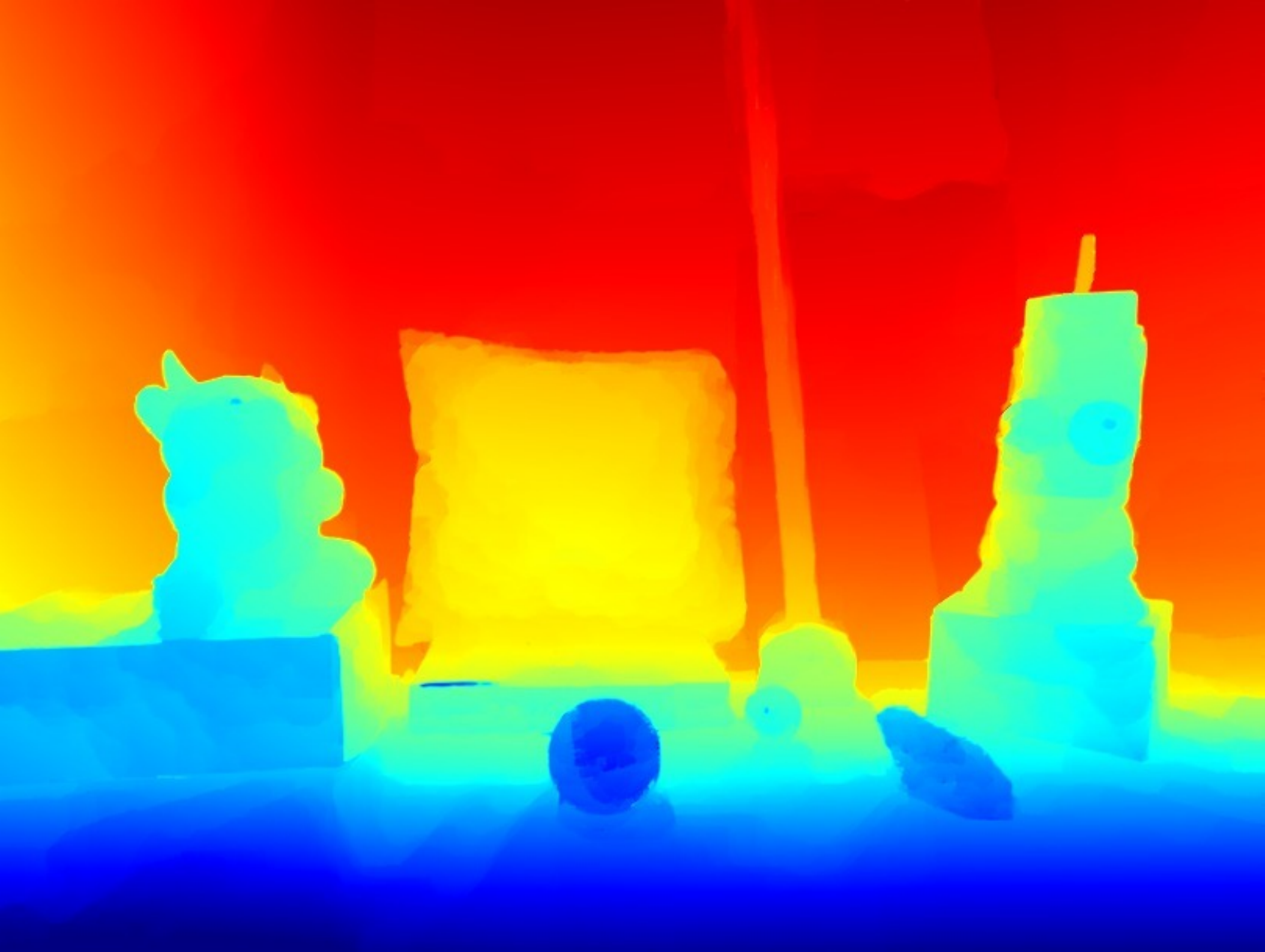}}\hspace*{0.1pt}
\includegraphics[height=\imglen]{colorbar_depth_with}\hspace*{0.0pt}
\subfloat[Error]{\includegraphics[height=\imglen]{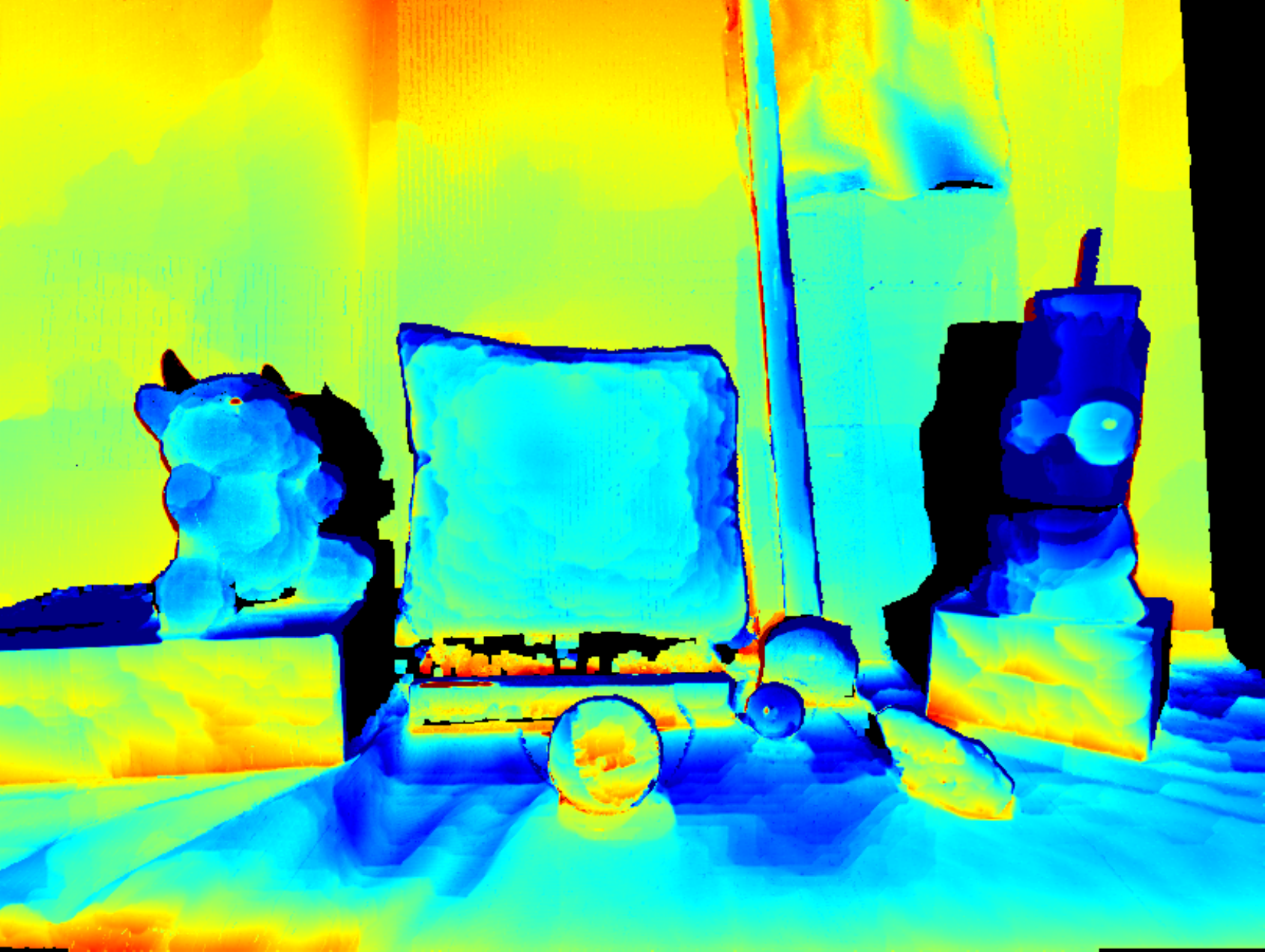}\hspace*{0.1pt}
\includegraphics[height=\imglen]{colorbar_error}}
\end{center}
  \caption{Visual evaluation on the real datasets \textit{Books} (first row), \textit{Shark} (second row) and \textit{Devil} (third row). In column (a) the low resolution \textit{ToF} image and the high resolution intensity image are shown, whereas column (b) shows the high resolution groundtruth depth. The black areas are not correctly reconstructed due to occlusions in the stereo system and therefore set invalid for the \textit{RMSE} calculation. In column (c) the upsampling result of our method is shown whereas in column (d) the relative depth error to the known groundtruth is shown.}
\label{fig:exp_real}
\end{figure*}

A quantitative accuracy evaluation of our \SR for three real world datasets is shown in Table \ref{tab:exp_real}. The upsampling error is calculated by the \RMSE to the groundtruth depth map measured with the highly accurate structured light scanner. We compared our method to two common interpolation techniques, joint bilateral upsampling \cite{Kopf2007} and guided image filtering \cite{He2010}. As depth input to all methods we used the offset corrected \TOF depth input. The average upsampling runtime of $318.2 \unit{ms}$ is measured as average over $100$ runs. The visual results are shown in Figure \ref{fig:exp_real}.

\begin{table}
  \begin{small}
  \begin{center}
    \begin{tabular}{@{}l*{4}{c}@{}}
      \toprule
       & \textit{Books} & \textit{Shark} & \textit{Devil} \\ \midrule 
      \RMSE nearest [$\unit{mm}$]  & 18.21 & 21.83 & 19.36 \\ 
      \RMSE bilinear [$\unit{mm}$] & 17.10 & 20.17 & 18.66 \\ 
      \RMSE Kopf et al. \cite{Kopf2007} [$\unit{mm}$]  & 16.03 & 18.79 & 27.57 \\ 
      \RMSE He et al. \cite{He2010} [$\unit{mm}$]  & 15.74 & 18.21 & 27.04 \\  \midrule
      Input depth density [$\unit{\%}$]  & 2.57  & 2.55  & 2.53 \\
      \RMSE \textit{OURS} [$\unit{mm}$] & 12.36 & 15.29 & 14.68 \\ \bottomrule
    \end{tabular}
  \end{center}
  \end{small}
  \caption{Quantitative evaluation on the real datasets \textit{Books}, \textit{Shark} and \textit{Devil}. The error is calculated as \textit{RMSE} to the measured groundtruth in $\unit{mm}$. We compare standard interpolation methods as well as joint bilateral filtering \cite{Kopf2007} and image guided filtering \cite{He2010} to our approach. The input density value shows the percentage of sparse depth values which are projected into the high resolution image space. This corresponds to an upsampling factor of approximately $\times 6.25$.}
\label{tab:exp_real}
\end{table}

\paragraph*{Discussion}
One issue that occurs in real world datasets is that wrong \TOF measurements result in displaced surfaces in the upsampled result. Another problem arises due to the difference in the viewpoint of the observing cameras. Thus, the projected depth measurements near large depth steps can differ from correct depth values. Because the distance between the cameras is very small compared to the measured depth range, these wrong measurements have no large impact on the result and can be handled by the regularization term. Despite that, in the visual and numerical results it can be seen that our method delivers high quality upsampling results at multiple frames per second for an approximate upsampling factor of $\times 6.25$. Through the additional incorporation of the \TOF sensor characteristics into our optimization, the acquisition noise is drastically reduced, while sharp edges and smooth surfaces are preserved. Compared to other methods it delivers superior results. Comparison to more \SOTA 
methods was not possible due to the lack of publicly available implementations, yet we provide a benchmarking framework to help others in benchmarking their methods.

\ResetAbbrevs{All}
\section{Conclusion}
In this paper we propose a novel method for depth map \SR using a low resolution, low cost 3D sensor and an additional high resolution 2D sensor. The \SR is formulated as a global energy optimization problem using \TGV regularization. For fast numerical optimization we use a first order primal-dual algorithm, which is efficiently parallelized resulting in high frame rates. In a quantitative evaluation using widespread datasets we show that our method clearly outperforms existing \SOTA methods in terms of speed and quality. We further provide benchmarking datasets of real world scenes providing a highly accurate groundtruth that, for the first time, enable a real quality comparison of depth image \SR methods. On these datasets we show a visual as well as numerical evaluation of our method.

The proposed method is not limited to single image upsampling. As a future perspective, it will be extended to incorporate a temporal coherence in a consistent way, eventually leading to depth reconstructions with even higher accuracy.


\ificcvfinal
\subsection*{Acknowledgments}
This work was supported by \textit{Infineon Technologies Austria AG} and the Austrian Research Promotion Agency (FFG) under the \textit{FIT-IT Bridge} program, project \#838513 (TOFUSION). 
\fi
{\small
\bibliographystyle{ieee}
\bibliography{egbib}
}

\end{document}